\documentclass[manuscript]{acmart}
\AtBeginDocument{%
  }

\setcopyright{acmlicensed}
\copyrightyear{2025}
\acmYear{2025}
\acmDOI{XXXXXXX.XXXXXXX}

\acmJournal{POMACS}
\acmVolume{37}
\acmNumber{4}
\acmArticle{111}
\acmMonth{8}

\usepackage{url}
\usepackage{times}
\usepackage{epsfig}
\usepackage{wrapfig}
\usepackage{graphicx}
\usepackage{amsmath}

\usepackage{amssymb}
\usepackage{multirow}
\usepackage{colortbl}
\usepackage{color}
\usepackage{threeparttable}
\usepackage{booktabs}
\usepackage{adjustbox}
\usepackage{enumitem}
\usepackage{subfig}
\usepackage{caption}
\usepackage[misc]{ifsym}
\usepackage{xcolor,colortbl}
\usepackage{makecell}
\usepackage{tikz}
\usepackage[edges]{forest}
\definecolor{hidden-draw}{RGB}{20,68,106}
\definecolor{hidden-pink}{RGB}{255,245,247}
\usepackage{placeins}
\usepackage[T1]{fontenc}

\usepackage[normalem]{ulem}

\usepackage{cancel}

\usepackage{hyperref}
\usepackage{url}
\usepackage{color}
\usepackage{colortbl}
\usepackage{soul}
\usepackage{multirow}

\usepackage{rotating,siunitx} %
\usepackage{tabularx}
\usepackage{hyphenat}
\usepackage{longtable}

\usepackage{float}

\usepackage{etoolbox}
\usepackage{algorithm}      
\usepackage[noend]{algpseudocode} 

\usepackage{pifont}

\newcommand{\encircle}[2][fill=lightcoral, text=white]{%
  \tikz[baseline=(char.base)]{%
    \node[circle, draw=none, thick, inner sep=1pt, #1] (char) {\textcolor{white}{#2}};
  }%
}

\newcommand{\eg}{\emph{e.g.,}}
\newcommand{\ie}{\emph{i.e.,}}
\newcommand{\eq}{Eq.}

\definecolor{lightcoral}{rgb}{0.94, 0.5, 0.5}
\definecolor{lightgreen}{rgb}{0.56, 0.93, 0.56}
\definecolor{harvestgold}{rgb}{0.85, 0.57, 0.0}
\definecolor{brightlavender}{rgb}{0.75, 0.58, 0.89}
\definecolor{capri}{rgb}{0.0, 0.75, 1.0}
\definecolor{carminepink}{rgb}{0.92, 0.3, 0.26}
\definecolor{celadon}{rgb}{0.67, 0.88, 0.69}
\definecolor{darkpastelgreen}{rgb}{0.01, 0.75, 0.24}

\definecolor{deepred}{rgb}{0.698,0.133,0.133}
\definecolor{blue}{rgb}{0,0,1}

\begin{document}

\title{Continual Learning for Generative AI: From LLMs to MLLMs and Beyond}

\author{Haiyang Guo}
\email{guohaiyang2023@ia.ac.cn}
\affiliation{%
  \institution{School of Advanced Interdisciplinary Sciences, University of Chinese Academy of Sciences}
  \city{Beijing}
  \country{China}
}

\author{Fanhu Zeng}
\affiliation{%
  \institution{MAIS, Institute of Automation, Chinese Academy of Sciences}
  \city{Beijing}
  \country{China}}
\email{zengfanhu2022@ia.ac.cn}

\author{Fei Zhu}
\affiliation{%
  \institution{Centre for Artificial Intelligence and Robotics, Hong Kong Institute of Science and Innovation, Chinese Academy of Sciences}
  \city{Hong Kong}
  \country{China}
}
\email{zhfei2018@gmail.com}

\author{Jiayi Wang}
\affiliation{%
  \institution{School of Advanced Interdisciplinary Sciences, University of Chinese Academy of Sciences}
  \city{Beijing}
  \country{China}
}

\author{Xukai Wang}
\affiliation{%
  \institution{MAIS, Institute of Automation, Chinese Academy of Sciences}
  \city{Beijing}
  \country{China}
}

\author{Jingang Zhou}
\affiliation{%
  \institution{School of Advanced Interdisciplinary Sciences, University of Chinese Academy of Sciences}
  \city{Beijing}
  \country{China}
}

\author{Hongbo Zhao}
\affiliation{%
  \institution{MAIS, Institute of Automation, Chinese Academy of Sciences}
  \city{Beijing}
  \country{China}
}

\author{Wenzhuo Liu}
\affiliation{%
  \institution{MAIS, Institute of Automation, Chinese Academy of Sciences}
  \city{Beijing}
  \country{China}
}

\author{Shijie Ma}
\affiliation{%
  \institution{MAIS, Institute of Automation, Chinese Academy of Sciences}
  \city{Beijing}
  \country{China}
}

\author{Da-Han Wang}
\affiliation{%
  \institution{School of Computer and Information Engineering, Xiamen University of Technology}
  \city{Xiamen}
  \country{China}
}

\author{Xu-Yao Zhang}
\authornote{This author is the corresponding author.}
\email{xyz@nlpr.ia.ac.cn}
\affiliation{%
  \institution{MAIS, Institute of Automation, Chinese Academy of Sciences}
  \city{Beijing}
  \country{China}
}

\author{Cheng-Lin Liu}
\email{liucl@nlpr.ia.ac.cn}
\affiliation{%
  \institution{MAIS, Institute of Automation, Chinese Academy of Sciences}
  \city{Beijing}
  \country{China}
}

\renewcommand{\shortauthors}{H. Guo et al.}

\begin{abstract}
  The rapid advancement of generative models has empowered modern AI systems to comprehend and produce highly sophisticated content, even achieving human-level performance in specific domains. However, these models are fundamentally constrained by \emph{catastrophic forgetting}, \ie~a persistent challenge where models experience performance degradation on previously learned tasks when adapting to new tasks. To address this practical limitation, numerous approaches have been proposed to enhance the adaptability and scalability of generative AI in real-world applications. In this work, we present a comprehensive survey of continual learning methods for mainstream generative AI models, encompassing large language models, multimodal large language models, vision-language-action models, and diffusion models. Drawing inspiration from the memory mechanisms of the human brain, we systematically categorize these approaches into three paradigms: architecture-based, regularization-based, and replay-based methods, while elucidating their underlying methodologies and motivations. We further analyze continual learning setups for different generative models, including training objectives, benchmarks, and core backbones, thereby providing deeper insights into the field. The project page of this paper is available at \href{https://github.com/Ghy0501/Awesome-Continual-Learning-in-Generative-Models}{https://github.com/Ghy0501/Awesome-Continual-Learning-in-Generative-Models}.
\end{abstract}





\maketitle

\section{Introduction}

The evolution of artificial intelligence is undergoing a paradigm shift from ``understanding the world'' to ``creating the world''. Traditional discriminative models, such as classification networks~\cite{krizhevsky2012imagenet, simonyan2014very, he2016deep, vaswani2017attention, zeng2024m2m, dosovitskiy2020image} and object detectors~\cite{girshick2014rich, ren2015faster, he2017mask, lin2017feature, redmon2016you}, have achieved remarkable success over the past decade. These models are competent in learning decision boundaries from large-scale labeled datasets, enabling accurate recognition and judgment of known concepts. However, this intelligence, based on discrimination, has shown limitations in creativity and generation. It excels at distinguishing known concepts within a single modality but struggles both to generate novel content and to establish semantic associations across multiple modalities. The rise of generative AI addresses these limitations by employing large-scale pre-training to capture the underlying data distribution. As a result, mainstream generative AI models such as Large Language Models~(LLMs)~\cite{achiam2023gpt, touvron2023llama, team2024gemini}, Multimodal Large Language Models~(MLLMs)~\cite{liu2023visual, chen2023minigpt, Qwen-VL, liu2024improved, bai2025qwen2}, Vision-Language-Action~(VLA) Models~\cite{kim2024openvla, zhou2025chatvla}, and Diffusion Models~\cite{rombach2022high, ho2020denoising} are now capable of not only interpreting input features but also actively generating coherent text, realistic images, and even cross-modal content. As the field transitions from discriminative to generative paradigms, AI is evolving from merely recognizing the known to actively creating the new. This shift is redefining the foundations of human-machine collaboration by enabling more flexible and multimodal generation capabilities.

\begin{figure*}[t]
    \centering
    \includegraphics[width=0.95\linewidth]{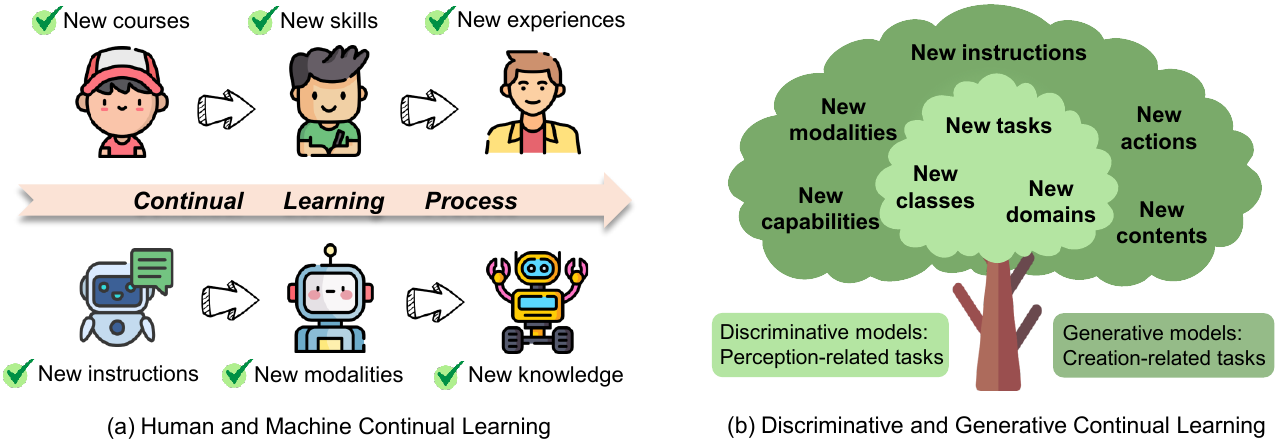}
    \caption{(a) Continual learning lets humans learn lifelong without forgetting; AI models aim to do the same while retaining old knowledge. (b) Discriminative models focus on perceptual tasks such as classification, detection, and segmentation, among others, whereas generative models additionally demand the capability to produce novel content grounded in perceptual understanding.}
    \label{fig:figure1}
    \vspace{-10pt}
\end{figure*}

The success of generative models like GPT-4~\cite{achiam2023gpt} fundamentally relies on large-scale generative pre-training and human alignment. However, this process is inherently dynamic, as languages, data distributions, and user demands continually evolve. More importantly, it is infeasible to account for all possible future scenarios in advance, which underscores the need for models to possess continual learning capabilities comparable to those of humans, as illustrated in Figure~\ref{fig:figure1}~(a). Such capabilities are therefore essential for generative models to move beyond the closed-world assumption~\cite{zhang2020towards} and enable deployment in open-ended, real-world scenarios~\cite{zhou2022open, parmar2023open, zhu2024open}. A central challenge in this context is \emph{catastrophic forgetting}, where learning new information causes previously acquired knowledge to be overwritten~\cite{mccloskey1989catastrophic, mcclelland1995there, kirkpatrick2017overcoming}. For instance, fine-tuning general-domain models (\eg~ChatGPT) on specialized domains (like mathematics) may improve in-domain performance at the expense of forgetting prior capabilities. When models are trained sequentially on multiple downstream tasks, each new update compounds this effect, leading to a progressive degradation in knowledge retention. A straightforward solution is to retain data from previous tasks and perform joint training with new data. However, this approach is often impractical due to significant storage costs~\cite{krempl2014open} and data privacy concerns~\cite{de2021continual}. By contrast, children and adults typically do not exhibit catastrophic forgetting when learning new information, even in the absence of explicit recall of prior knowledge~\cite{xue2010greater, chang2025sleep}, highlighting a fundamental gap between artificial and biological learning systems. To bridge this gap, the study of continual learning~\cite{van2022three, li2017learning, lopez2017gradient, rolnick2019experience, zhu2021prototype} has gained momentum, aiming to equip models with the ability to learn new knowledge over time while retaining prior skills, especially in the practical cases where the data from old tasks are limited or even inaccessible, particularly in practical scenarios where data from earlier tasks is limited or entirely unavailable.

\begin{figure*}[t]
    \centering
    \includegraphics[width=0.98\linewidth]{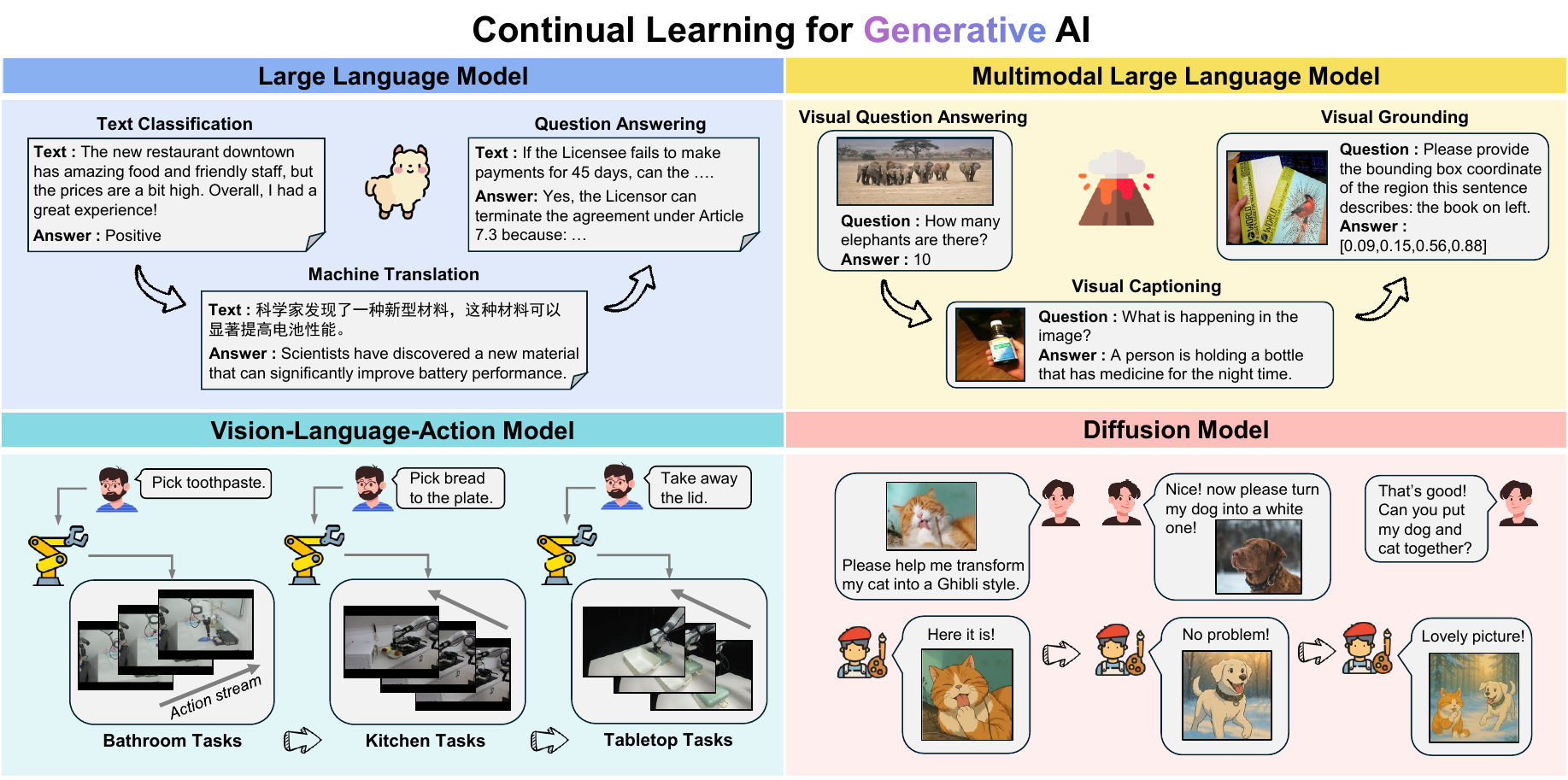}
    \caption{Illustrations of continual learning for generative AI models: Adapting to real-world demands by learning new instructions, knowledge, modalities, actions, and creations without catastrophic forgetting.}
    \label{fig:figure2}
    \vspace{-10pt}
\end{figure*}

Compared to traditional discriminative models that focus on perceptual tasks, generative models in continual learning settings are expected to go a step further, as shown in Figure~\ref{fig:figure1}~(b), by creating new content grounded in understanding and recognition. We summarize their increased challenges in continual learning into two main aspects:~(1) More complex modeling objectives. Discriminative methods focus on modeling posterior and discriminative probabilities~(\ie~$P(y|x)$), where catastrophic forgetting primarily manifests as a drift in the decision boundary within the classification space~\cite{zhu2025pass_plus, zhou2024expandable, liu2025class, zhang2023slca, guo2024desire, zhu2021class}. In contrast, generative models aim to grasp the underlying probability distribution~(\ie~$P(x)$), which requires continual learning to maintain both semantic coherence and knowledge completeness within an open-ended generative space. To achieve this, it is essential to refine the generative reasoning process alongside the underlying factual knowledge representation, thereby mitigating hallucinations~\cite{huang2025survey, bai2024hallucination} and semantic inconsistencies~\cite{zhang2024respond}.
(2)~Greater diversity in task formats. Generative models encounter a wide variety of tasks during continual learning, with differences arising from factors such as new input modalities, novel domain knowledge, and evolving task characteristics. This variability increases the risk of catastrophic forgetting and places greater demands on multi-task knowledge integration. In contrast, discriminative models typically focus on a narrow range of tasks, such as classification, detection, or segmentation, which makes continual learning comparatively less challenging. Figure~\ref{fig:figure2} demonstrates the continual learning scenarios across four representative generative models. Despite their impressive capabilities, enabling these models to learn continuously remains challenging and underexplored.

Several surveys focus on addressing continual learning and catastrophic forgetting, but they exhibit certain limitations. First, reviews focusing on discriminative models have attracted significant interest~\cite{masana2022class, wang2024comprehensive, zhou2024class,menezes2023continual, yuan2024survey}, but due to fundamental differences between discriminative and generative paradigms, these methods cannot be straightforwardly applied to continual learning in generative models. Furthermore, existing surveys on continual learning for generative models generally concentrate on specific architectures, such as LLMs~\cite{wu2024continual, shi2024continual, zheng2025towards, yang2025recent} or MLLMs~\cite{yu2024recent, huo2025continue, huang2025keeping}. Although these works provide systematic overviews within their respective domains, they often lack comprehensive discussions on the relationships, similarities, and differences among various types of generative models. To fill this gap, in this work, we present the first comprehensive review of continual learning across a wide spectrum of mainstream generative models, including LLM, MLLM, VLA, and Diffusion Model. Specifically, we start by uncovering the underlying principle in the context of continual learning shared across these types of models from a unified perspective. Subsequently, we systematically analyze the key approaches and challenges in each type of generative model, respectively. As a whole, our paper provides up-to-date advances in the literature and offers a holistic perspective for future works.

The paper is organized as follows: In Section~\ref{sec:overview}, we present the setup of continual learning, covering its basic formulation, evaluation metrics, and taxonomy. Sections~\ref{Sec:LLM} through~\ref{Sec:Diff} provide an in-depth discussion of continual learning in Large Language Models~(Section~\ref{Sec:LLM}), Multimodal Large Language Models~(Section~\ref{Sec:MLLM}), Vision-Language-Action Models~(Section~\ref{Sec:VLA}), and Diffusion Models~(Section~\ref{Sec:Diff}). Each domain is examined in terms of training strategies for continual learning tasks, representative benchmarks, model architectures, as well as the motivations and practical implementations of existing approaches. In Section~\ref{sec:future}, we further explore potential future research directions and emerging trends in continual learning methods built upon generative models. Finally, Section~\ref{sec:conclusion} concludes the paper and introduces a continuously updated \href{https://github.com/Ghy0501/Awesome-Continual-Learning-in-Generative-Models}{GitHub repository} for tracking the latest progress in continual learning for generative models.

\section{Overview of Continual Learning}
\label{sec:overview}

\subsection{Basic Formulation}

Continual learning tackles the challenge of acquiring knowledge from dynamically evolving data distributions, where training samples from distinct tasks arrive sequentially. A continual learning model is required to assimilate new knowledge while preserving previously learned information, typically under constraints that limit or prevent access to past data, while maintaining strong performance across all tasks. Formally, suppose there are $T$ continual learning tasks in total. The training samples for task $t$ can be represented as $\mathcal{D}_t=\{(x, y)\}$, where $x$ is the input data and $y$ is the corresponding label. Notably, the types of training data required vary across different generative models. For example, in MLLM, $(x, y)$  typically consists of a triplet comprising input images, user instructions, and target answers $(\textbf{X}_{v}, \textbf{X}_{ins}, \textbf{X}_{a})$. In VLA models, it can take the form of state-action pairs $(\mathcal{S}, \mathcal{A})$, where the state $\mathcal{S}$ includes an image observation and a text instruction. During the training of task $t$, only the data for the current task is accessible. The training objective of each continual learning task can be formalized as follows:
\begin{equation}
    \mathop{\arg\min}_{\theta}~\mathbb{E}_{(x, y)\sim\mathcal{D}_t}\left[\ell(f_{\theta}(x),y))\right],
\label{eq:eq1}
\end{equation}
where $f_{\boldsymbol{\theta}}$ represents the generative model, and $\ell(\cdot, \cdot)$ is the task-specific loss function. The specific formulation for each generative model is detailed at the beginning of its corresponding section. 

Upon completion of training for the $t$-th task, the model is evaluated on the test sets of all previously seen tasks, without having access to any task identity information for the current input. An ideal continual learning framework for generative models should enable comprehensive learning of the current task, preserve previously acquired knowledge, and demonstrate strong generalization to future unseen tasks.

\subsection{Evaluation Metrics}
The evaluation of a model’s continual learning ability typically involves three aspects: overall performance on all learned tasks, the extent of forgetting on previously learned tasks, and the generalization capability to unseen tasks. For clarity, we denote $a_{t,k}$ as the performance on the $k$-th task after the model has been sequentially trained from task 1 to task $t$~\cite{wang2024comprehensive}.

\textbf{Overall performance} typically includes \emph{Last Accuracy}~(Last) and \emph{Average Accuracy}~(Avg). These two metrics are calculated after training on the $t$-th task as follows:
\begin{equation}
    \text{Avg}_t = \frac{1}{t}\sum_{j=1}^t \text{Last}_j,~\text{Last}_t = \frac{1}{t}\sum_{i=1}^t a_{t,i},
\end{equation}
\noindent
where Last reflects the model's average performance on the current task and Avg further synthesizes the performance across previous tasks. For both metrics, higher values correspond to superior continual learning performance.

\textbf{Forgetting evaluation} typically includes two key metrics: the \emph{forgetting measure}~(FM) and \emph{backward transfer}~(BWT). The former first computes the forgetting degree for each seen task, derived from the difference between the historically highest accuracy and the current accuracy:
\begin{equation}
    f_{k,t}=\max_{i \in \{1,\cdots,t-1\}} (a_{i,k}-a_{t,k}), \forall k<t.
\end{equation}
FM is calculated as the average forgetting degree across all previous tasks at task $t$:
\begin{equation}
    \text{FM}_t=\frac{1}{t-1}\sum_{i=1}^{t-1} f_{i,t}.
\end{equation}
BWT quantifies the average impact of learning the $t$-th task on performance across all previously learned tasks:
\begin{equation}
    \text{BWT}_t=\frac{1}{t-1}\sum_{k=1}^{t-1} (a_{t,k}-a_{k,k}).
\end{equation}
FM quantifies catastrophic forgetting, where lower values indicate better retention of previous knowledge. BWT evaluates knowledge transfer between tasks, with positive values (\emph{i.e.}~BWT > 0) reflecting beneficial forward transfer to prior tasks and negative values (\emph{i.e.}~BWT < 0) indicating interference.

\textbf{Generalization capability} can be evaluated by \emph{zero-shot transfer}~(ZT), which is designed to quantify how well zero-shot transferability is maintained throughout the training process. Formally, for task $k$, it calculates the average accuracy $A_k$ of models trained on preceding tasks evaluated on task $k$:
\begin{equation}
    A_{k} = \frac{1}{k-1}\sum_{i=1}^{k-1}a_{i,k}, \forall k>1.
\end{equation}
ZT is the average of $A_k$ after training on the final task:
\begin{equation}
    \text{ZT}=\frac{1}{T-1}\sum_{k=2}^{T}A_k.
\end{equation}
A higher value of ZT demonstrates superior preservation of zero-shot generalization capability for unseen tasks throughout the continual learning process.

It is worth noting that the calculation of $a_{t,k}$ depends on the specific task type. For example, Bilingual Evaluation Understudy~(BLEU)~\cite{papineni2002bleu} for text generation task, Intersection-over-Union (IoU)~\cite{cermelli2020modeling} for visual grounding task, Fr\'echet Inception Distance (FID)~\cite{heusel2017gans} for text-to-image generation task, and so on. There are several other evaluation criteria available, and we refer readers to their original papers for detailed specifications.

\begin{figure*}[t]
    \centering
    \tikzset{
            my node/.style={
                draw,
                align=center,
                thin,
                text width=2.5cm, 
                rounded corners=3,
            },
            my leaf/.style={
                draw,
                align=left,
                thin,
                text width=4.5cm, 
                rounded corners=3,
            }
    }
    \forestset{
      every leaf node/.style={
        if n children=0{#1}{}
      },
      every tree node/.style={
        if n children=0{minimum width=1em}{#1}
      },
    }
    \begin{forest}
      nonleaf/.style={font=\bfseries\scriptsize},
           for tree={%
              every leaf node={my leaf, font=\scriptsize},
              every tree node={my node, font=\scriptsize, l sep-=4.5pt, l-=1.pt},
              anchor=west,
              inner sep=2pt,
              l sep=10pt, %
              s sep=3pt, %
              fit=tight,
              grow'=east,
              edge={ultra thin},
              parent anchor=east,
              child anchor=west,
              ver/.style={rotate=90, child anchor=north, parent anchor=south, anchor=center},
              if n children=0{}{nonleaf}, 
              edge path={
                  \noexpand\path [draw, \forestoption{edge}] (!u.parent anchor) -- +(5pt,0) |- (.child anchor)\forestoption{edge label};
              },
              if={isodd(n_children())}{
                  for children={
                      if={equal(n,(n_children("!u")+1)/2)}{calign with current}{}
                  }
              }{}
          }
      [\textbf{Continual Learning for Generative AI}, draw=gray, color=gray!100, fill=gray!15, very thick, text=black, text width=4cm, ver, font=\normalsize
        [Large Language Models \\ (\S~\ref{Sec:LLM}), color=lightcoral!100, fill=lightcoral!15, very thick, text=black, text width=2.5cm
            [Architecture-based \\ (\S~\ref{LLM:Architecture-based}), color=lightcoral!100, fill=lightcoral!15, very thick, text=black, text width=2.5cm
                [{TPEM~\cite{geng2021continual}, Zheng et al~\cite{zheng2025spurious}, TGL~\cite{fernandez2024gradient}, LOIRE~\cite{hanloire}, AdapterCL~\cite{madotto2020continual}, SAPT~\cite{zhao2024sapt}, MoRAL~\cite{yang2024moral}, I-LoRA~\cite{ren2024analyzing}, SLIM~\cite{han2024slim}, SEE~\cite{wang2025see}, \\ P-Prompts~\cite{razdaibiedina2023progressive},  Q-Tuning~\cite{guo2024q}, WAVE-CRE~\cite{le2025adaptive}
                }, color=lightcoral!100, fill=lightcoral!15, very thick, text=black, text width=7.25cm]
                ]
            [Regularization-based \\ (\S~\ref{LLM:Regularization-based}), color=lightcoral!100, fill=lightcoral!15, very thick, text=black, text width=2.5cm
                [{O-LoRA~\cite{wang2023orthogonal}, DAS~\cite{ke2023continual}, ARPER~\cite{mi2020continual}, SEEKR~\cite{he2024seekr}, TaSL~\cite{feng2024tasl}, FVG~\cite{jiang2025unlocking}, \\
                DaCL~\cite{wu2023enhancing}, DYNAINST~\cite{mok2023large},  
                Velocitune~\cite{luo2024velocitune},
                Recurrent-KIF~\cite{feng2025recurrent}}, color=lightcoral!100, fill=lightcoral!15, very thick, text=black, text width=7.25cm]
                ]
            [Replay-based \\ (\S~\ref{LLM:Replay-based}), color=lightcoral!100, fill=lightcoral!15, very thick, text=black, text width=2.5cm
                [{CT0~\cite{scialom2022fine}, Huang et al.~\cite{huang2024towards}, Anh et al.~\cite{anh2025mutual}, InsCL~\cite{wang2024inscl}, HESIT~\cite{chen2024overcoming}, \\ Sun et al.~\cite{sun2024reviving}, D-CPT Law~\cite{que2024d}, LAMOL~\cite{sun2019lamol}, HMI-LAMOL~\cite{maekawa2023generative}, PCLL~\cite{zhao2022prompt},  DCL~\cite{zeng2023continual}, SSR~\cite{huang2024mitigating},
                SynE~\cite{chen2024towards},
                KPIG~\cite{he2024don},
                Guo et al.~\cite{guo2024efficient}
                }, color=lightcoral!100, fill=lightcoral!15, very thick, text=black, text width=7.25cm]
            ]
            ]
        [Multimodal Large Language Models \\ (\S~\ref{Sec:MLLM}), color=harvestgold!100, fill=harvestgold!15, very thick, text=black, text width=2.5cm
            [Architecture-based \\ (\S~\ref{MLLM:art}), color=harvestgold!100, fill=harvestgold!15, very thick, text=black, text width=2.5cm
                [{Eproj~\cite{he2023continual},
                TAM-CL~\cite{cai2023task}, MoELoRA~\cite{chen2024coin}, Continual LLaVA~\cite{cao2024continual}, HiDe‑LLaVA~\cite{guo2025hide}, LLaCA~\cite{qiao2024llaca}, SMoLoRA~\cite{wang2024separable}, CL-MoE~\cite{huai2025cl}, \\ LLaVA-CMoE~\cite{zhao2025llava}, DISCO~\cite{guo2025federated}, TRIPLET~\cite{qian2023decouple},  ModalPrompt~\cite{zeng2024modalprompt}, \\  Fwd-Prompt~\cite{zheng2024beyond},  ColPro~\cite{cai2024empowering}, CluMo~\cite{cai2024clumo}, 
                MR-LoRA~\cite{hongbozhao25}, 
                D-MoLE~\cite{ge2025dynamic},
                BranchLoRA~\cite{zhang2025enhancing},
                ProgLoRA~\cite{yu2025progressive}
                }, color=harvestgold!100, fill=harvestgold!15, very thick, text=black, text width=7.25cm]
            ]
            [Regularization-based \\ (\S~\ref{MLLM:reg}), color=harvestgold!100, fill=harvestgold!15, very thick, text=black, text width=2.5cm
                [{SCD~\cite{lao2023multi}, CS-VQLA~\cite{bai2023revisiting}, Chen et al.~\cite{chen2024llm}, CrossSDC~\cite{pian2024continual},  MAFED~\cite{nikandrou2024enhancing}, \\ MoInCL~\cite{pian2024modality}, QUAD~\cite{marouf2025no}, Model Tailor~\cite{zhu2024model}, SPIDER~\cite{huang2024learn},
                SEFE~\cite{chen2025sefe}, \\ 
                LoRASculpt~\cite{liang2025lorasculpt},
                LLaVA-c~\cite{liu2025llava}
                }, color=harvestgold!100, fill=harvestgold!15, very thick, text=black, text width=7.25cm]
            ]
            [Replay-based \\ (\S~\ref{MLLM:replay}), color=harvestgold!100, fill=harvestgold!15, very thick, text=black, text width=2.5cm
                [{VQACL~\cite{zhang2023vqacl}, SGP~\cite{lei2023symbolic}, ProtoGroup~\cite{zhang2025multi}, Lin et al.~\cite{lin2025vlm}, Adapt-$\infty$~\cite{maharana2024adapt}, \\ OASIS~\cite{lee2025oasis}}, color=harvestgold!100, fill=harvestgold!15, very thick, text=black, text width=7.25cm]
            ]
            ]
        [Vision-Language-Action \\ Models \\ (\S~\ref{Sec:VLA}), color=cyan!100, fill=cyan!15, very thick, text=black, text width=2.5cm
            [Architecture-based \\ (\S~\ref{sec:vla-arch}), color=cyan!100, fill=cyan!15, very thick, text=black, text width=2.5cm
                [{LIBERO~\cite{liu2023libero}, Lotus~\cite{wan2024lotus}, QueST~\cite{mete2024quest}, LEGION~\cite{meng2025preserving},
                Jia et al.~\cite{jia2025hierarchical}, \\
                DRAE~\cite{long2025drae}
                }, color=cyan!100, fill=cyan!15, very thick, text=black, text width=7.25cm]
            ]
            [Regularization-based \\ (\S~\ref{sec:vla-reg}), color=cyan!100, fill=cyan!15, very thick, text=black, text width=2.5cm
                [{CAMA~\cite{kim2024online}, M2Distill~\cite{roy2024m2distill}}, color=cyan!100, fill=cyan!15, very thick, text=black, text width=7.25cm]
            ]
            [Replay-based \\ (\S~\ref{sec:vla-replay}), color=cyan!100, fill=cyan!15, very thick, text=black, text width=2.5cm
                [{iManip~\cite{zheng2025imanip}, RWLA~\cite{yang2024task}}, color=cyan!100, fill=cyan!15, very thick, text=black, text width=7.25cm]
            ]
            ]
        [Diffusion Models \\ (\S~\ref{Sec:Diff}), color=darkpastelgreen!100, fill=darkpastelgreen!15, very thick, text=black, text width=2.5cm
            [Architecture-based \\ (\S~\ref{Diff:arch}), color=darkpastelgreen!100, fill=darkpastelgreen!15, very thick, text=black, text width=2.5cm
                [{C-LoRA~\cite{smith2023continual}}, color=darkpastelgreen!100, fill=darkpastelgreen!15, very thick, text=black, text width=7.25cm]
            ]
            [Regularization-based \\ (\S~\ref{Diff:reg}), color=darkpastelgreen!100, fill=darkpastelgreen!15, very thick, text=black, text width=2.5cm
                [{MuseumMaker~\cite{liu2024museummaker}, LFS-Diffusion~\cite{song2024towards}, CIDM~\cite{dong2024continually}, ConceptGuard~\cite{guo2025conceptguard}, \\
                Jha et al~\cite{jha2025mining}}, color=darkpastelgreen!100, fill=darkpastelgreen!15, very thick, text=black, text width=7.25cm]
            ]
            [Replay-based \\ (\S~\ref{Diff:replay}), color=darkpastelgreen!100, fill=darkpastelgreen!15, very thick, text=black, text width=2.5cm
                [{$\text{L}^2\text{DM}$~\cite{sun2024create}}, color=darkpastelgreen!100, fill=darkpastelgreen!15, very thick, text=black, text width=7.25cm]
            ]
            ]
      ]
    \end{forest}
\caption{Taxonomy of Continual Learning for Generative AI.}
\vspace{-10pt}
\label{fig:figure3}
\end{figure*}
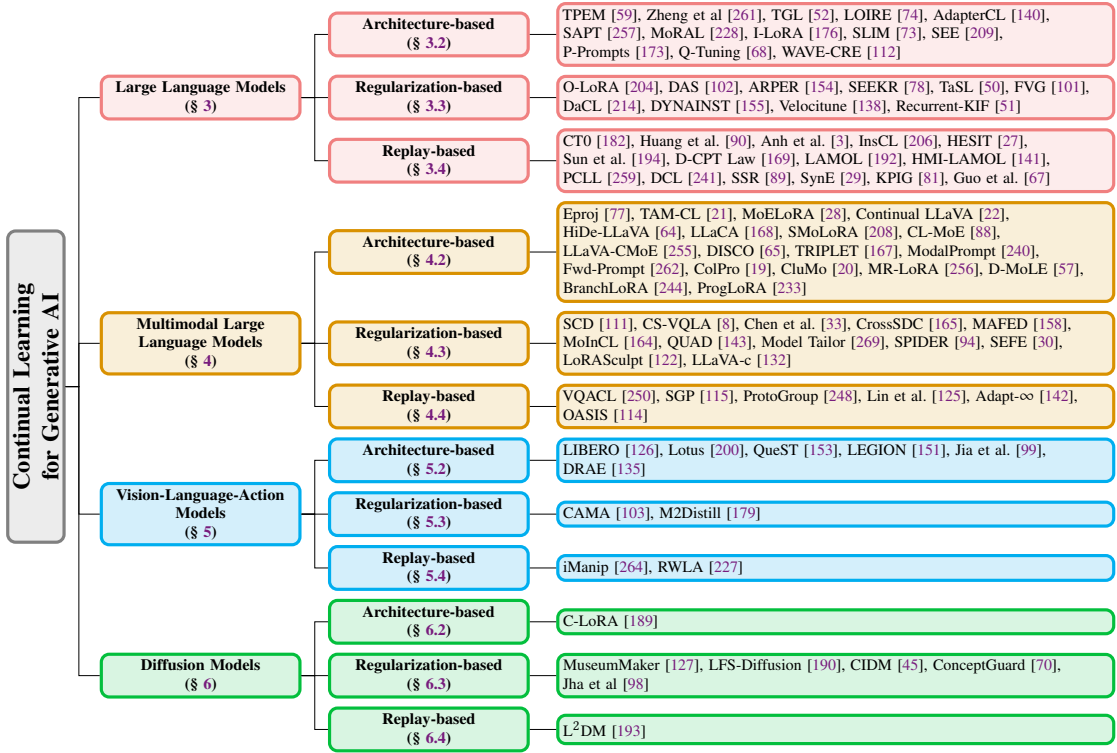

\subsection{Taxonomy}

Inspired by the collaborative mechanisms of rapid learning in the hippocampus and long-term integration in the neocortex~\cite{jeong2025goal, moscovitch2016episodic, sirota2003communication, kumaran2016learning, mcclelland1996considerations}, continual learning methods can be systematically categorized into the following three approaches.

\textbf{Architecture-based methods} mimic the brain's modular organization~\cite{aso2016dopaminergic, aso2014mushroom, frankland2005organization, euston2012role} through dynamic network expansion or modular design to isolate task-specific knowledge. These methods typically encapsulate such knowledge within distinct network modules. As model sizes grow, existing approaches increasingly adopt Parameter-Efficient Fine-Tuning~(PEFT) techniques~(\eg~LoRA~\cite{hu2022lora} and Prompts~\cite{lester2021power}) to store and adapt knowledge. Formally, let $\theta_{old}$ denote the model parameters that remain frozen during training on the current task. These typically include the backbone network and the extended sub-network associated with previous tasks. The parameters specific to the current task are denoted by $\theta_{new}$, which are usually the only trainable components. At this point, Eq~(\ref{eq:eq1}) is rewritten as follows:
\begin{equation}
    \mathop{\arg\min}_{\theta_{new}}~\mathbb{E}_{(x, y)\sim\mathcal{D}_t}\left[\ell(f_{\theta_{old}\bigcup\theta_{new}}(x),y))\right].
\end{equation}
Since the generative model inherently possesses strong generalization capabilities, these submodules with relatively few parameters can effectively retain task-specific knowledge. During inference, one of the key challenges for this class of methods is to effectively select the appropriate sub-module based on the input sample.

\textbf{Regularization-based methods} emulate neocortical synaptic stability and metaplasticity by constraining critical parameters to preserve learned representations, mirroring biological selective consolidation~\cite{hayashi2015labelling, yang2009stably, arevian2008activity, flesch2022orthogonal, xie2022geometry}. To achieve this goal, generative models often employ regularization in either the parameter space or the feature space, with the former formalized as follows:
\begin{equation}
    \mathop{\arg\min}_{\theta}~\mathbb{E}_{(x, y)\sim\mathcal{D}_t}\left[\ell(f_{\theta}(x),y)) + 
     \lambda \cdot \Omega(\theta, \theta_{old})\right],
\end{equation}
where $\lambda$ is a pre-defined hyperparameter, and $\Omega(\cdot, \cdot)$ is a regularization term~(\eg~L2 norm) enforcing constraints between the current parameter $\theta$ and the old task parameter $\theta_{old}$.
\begin{equation}
    \mathop{\arg\min}_{\theta}~\mathbb{E}_{(x, y)\sim\mathcal{D}_t}\left[\ell(f_{\theta}(x),y)) + 
     \lambda \cdot \mathcal{R}(\phi, \phi_{old})\right],
\end{equation}
where $\phi=f_{\theta}(x)$ denotes the output features or logits extracted by the model, and $\mathcal{R}(\cdot, \cdot)$ is a feature regularization term~(\eg~knowledge distillation~\cite{hinton2015distilling}) used to preserve the representations learned from previous tasks. The central challenge of such methods is to devise regularization terms that effectively retain knowledge from prior tasks with minimal impact on the acquisition of new knowledge.

\textbf{Replay-based methods} replicate hippocampal memory replay mechanisms~\cite{davidson2009hippocampal, gupta2010hippocampal, carr2011hippocampal, olafsdottir2018role} by storing raw data or generating synthetic samples to revisit past experiences during new task training, effectively mitigating catastrophic forgetting. In practice, this approach typically involves maintaining an auxiliary memory buffer $\mathcal{M}$ that stores a subset of data from previous tasks, which is then used in joint training with current task data:
\begin{equation}
    \mathop{\arg\min}_{\theta}~\mathbb{E}_{(x, y)\sim(\mathcal{D}_t\bigcup\mathcal{M})}\left[\ell(f_{\theta}(x),y))\right].
\end{equation}
For generative models, a key challenge in replay-based continual learning methods stems from the limited capacity of memory buffers. Given this constraint, selecting a small yet representative subset of past samples for rehearsal becomes a critical issue. Furthermore, considering practical concerns such as data privacy and storage overhead, several approaches opt to retain intermediate representations (\eg~features or hidden states) of previous tasks rather than storing raw samples, thereby mitigating the risks and costs associated with direct data retention.

Accordingly, we categorize existing generative model-based continual learning approaches according to the taxonomy outlined above, as illustrated in Figure~\ref{fig:figure3}. Notably, reflecting the brain's distributed processing of complex information via interconnected neural units, these methods often consist of multiple interdependent modules. For analytical clarity, each method is primarily classified based on its main functional component, while additional design dimensions are introduced in detail within the corresponding descriptive sections.

\section{Continual Learning for Large Language Models}
\label{Sec:LLM}
In recent years, Large Language Models~(LLMs)~\cite{achiam2023gpt,grattafiori2024llama,yang2024qwen2} have demonstrated outstanding natural language understanding and generation capabilities by pre-training on massive general-domain text corpora. To sustain high performance in real-world applications, LLMs require the capability to integrate new information and retain prior knowledge as data, tasks, and user preferences evolve. Unlike traditional language models, LLMs treat every task as a generative sequence prediction problem through autoregressive next-token prediction. Their massive parameterization enables emergent capabilities including few-shot learning~\cite{brown2020language} and complex multi-step reasoning~\cite{wei2022chain}. Consequently, many conventional continual learning methods designed for smaller models struggle to adapt effectively to LLMs.

In the following section, we first outline the setup for an LLM continual learning task~(Section \ref{LLM:setup}), including its training objectives, commonly used model backbones, and widely adopted benchmarks. Next, we classify the existing research into three strategies: architecture-based~(Section \ref{LLM:Architecture-based}), regularization-based~(Section \ref{LLM:Regularization-based}), and replay-based~(Section \ref{LLM:Replay-based}) approaches.

\subsection{Problem Setup}
\label{LLM:setup}

\textbf{Training objectives.} 
In the instruction tuning paradigm for a LLM $f_\theta$, the training dataset $\mathcal{D} = \{(x_j, y_j)\}_{j=1}^N$ typically consists of input texts $x$~(\eg~questions) paired with target output texts $y$~(\eg~answers), where $N$ denotes the total number of samples. Given an input $x$ and its corresponding output $y$ of length $L$, the model generates $y$ autoregressively:
\begin{equation}
    p(y \mid x) = \prod_{i=1}^L p_{\boldsymbol{\theta}}(y_i \mid x, y_{<i}),
\end{equation}
where $\theta$ denotes the trainable parameters during fine-tuning, $y_i$ is the $i$-th token in the output sequence, and $y_{<i}$ represents all previously generated tokens before position $i$. The objective of instruction tuning is to minimize the overall loss on the dataset $\mathcal{D}$:
\begin{equation}
    \min_{\boldsymbol{\theta}}\mathcal{L}(\boldsymbol{\theta})=\mathbb{E}_{(\boldsymbol{x},\boldsymbol{y})\sim\mathcal{D}}\left[\ell(f_{\boldsymbol{\theta}}(\boldsymbol{x}),\boldsymbol{y})\right],
\label{eq:llm-loss}
\end{equation}
where $x$ and $y$ denote the input and target texts, respectively, and $\ell(\cdot, \cdot)$ is the loss function, typically the token-level cross-entropy loss, which measures the discrepancy between the predicted output and the ground-truth target.

\begin{figure*}[t]
    \centering
    \includegraphics[width=0.98\linewidth]{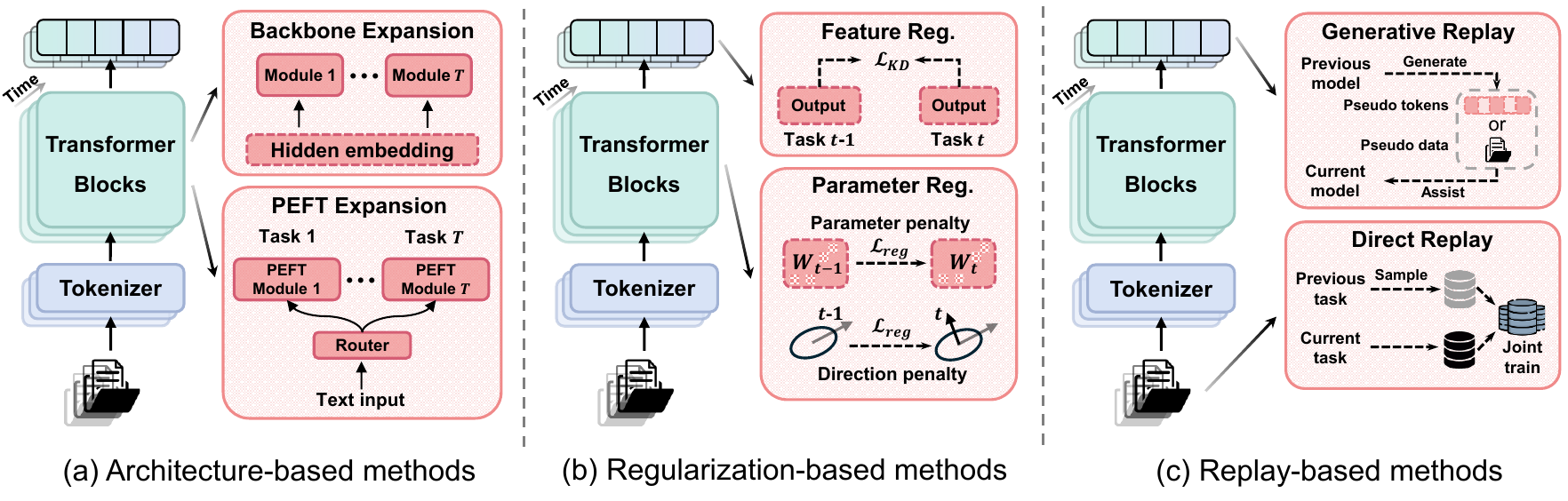}
    \vspace{-5pt}
    \caption{Overview of continual learning approaches in Large Language Models.}
    \label{fig:llm}
    \vspace{-10pt}
\end{figure*}


\noindent
\textbf{Backbones.}
Mainstream LLMs for natural language generation typically follow two architectures: encoder-decoder and decoder-only. In continual learning tasks, common backbones include:
\begin{itemize}
    \item \textbf{BERT}~\cite{devlin2019bert} employs a bidirectional encoder-only Transformer. Unlike autoregressive models, it uses masked language modeling and next-sentence prediction to learn contextualized representations from both directions. Each layer contains multi-head self-attention, feedforward networks, layer normalization, and residual connections.
    \item \textbf{T5}~\cite{raffel2020exploring} uses a standard encoder-decoder Transformer, unifying all NLP tasks as text-to-text generation. Its layers combine multi-head attention and feedforward sublayers, with configurable size depending on the model scale.
    \item \textbf{GPT}~\cite{radford2019gpt} series exemplify decoder-only architectures with left-to-right autoregressive generation. GPT stacks masked multi-head attention and feedforward layers with residual connections, optimized for open-ended generation. GPT-2 is often the preferred version in research.
    \item \textbf{LLaMA}~\cite{touvron2023llama} retains the decoder-only design but adds improvements like Rotary Positional Embeddings (RoPE), SwiGLU activation, and RMSNorm, enhancing training stability, inference speed, and long-context handling.
\end{itemize}

In addition to the mainstream backbone networks mentioned above, some approaches use other architectures as well, and we refer readers to their original papers.

\noindent
\textbf{Benchmarks.}
Several benchmarks evaluate the continual learning performance of LLMs. Key representative ones include:

\begin{itemize}
    \item \textbf{Short-sequence CL}~\cite{zhang2015character} consists of five large-scale text classification datasets: AG News (4 classes), Amazon Reviews (5 classes), Yelp Reviews (5 classes), DBpedia (14 classes), and Yahoo Answers (10 classes).
    \item \textbf{Long-sequence CL}~\cite{razdaibiedina2023progressive} extends Short-sequence CL by adding four GLUE tasks (MNLI, QQP, RTE, SST-2), five SuperGLUE tasks (WiC, CB, COPA, MultiRC, BoolQ), and IMDB sentiment analysis.
    \item \textbf{SuperNI}~\cite{wang2022super} is a large benchmark with 1,616 NLP tasks spanning 76 task types and 55 languages. Each task has detailed instructions and examples, supporting diverse continual learning with varied domain and task subsets.
    \item \textbf{TRACE}~\cite{wang2023trace} covers eight challenging datasets involving domain knowledge, multilinguality, code generation, and mathematical reasoning. It standardizes dataset formats and introduces three metrics for forgetting: (1) General Ability Delta (factual QA, reasoning, multilingual QA, commonsense, comprehension), (2) Instruction Following Delta (adherence to Self-instruct and LIMA prompts), and (3) Safety Delta (response safety via CoNa dataset).
\end{itemize}

\begin{table*}[!t]
    \centering
    \caption{Summary of continual learning methods for Large Language Models. Taxonomy: \textcolor{red}{\ding{52}} indicates the primary solution, and \ding{52} indicates the supporting method. Setting: CToD refers to Continual Task-oriented Dialogue, CIT to Continual Instruction Tuning, CPT to Continual Pre-Training, and CRE to Continual Relation Extraction. } 
    \resizebox{0.98\textwidth}{!}{
    \begin{tabular}{lcccclcc}
    \toprule
       \multirow{2}{*}{\textbf{Methods}} & \multicolumn{3}{c}{\textbf{Taxonomy}} & \multirow{2}{*}{\textbf{Setting}} & \multirow{2}{*}{\textbf{Backbone / Basic models}}  & \multirow{2}{*}{\textbf{Implement}} & \multirow{2}{*}{\textbf{Code}} \\ 
       \cmidrule{2-4}
       & \textbf{Arch.} & \textbf{Reg.} & \textbf{Rep.}  \\ \midrule
       \rowcolor[rgb]{ .949,  .949,  .949} TPEM~\cite{geng2021continual} & \textcolor{red}{\ding{52}} & \ding{52} &  & CToD & GLMP & Backbone & \href{https://github.com/siat-nlp/TPEM}{Link} \\ 
       Zheng et al.~\cite{zheng2025spurious} & \textcolor{red}{\ding{52}} &  &  & CIT & LLaMA-2-7B-Chat, LLaMA-3-8B-Instruct, Pythia-410M & Backbone & \href{https://github.com/zzz47zzz/spurious-forgetting}{Link} \\ 
       \rowcolor[rgb]{ .949,  .949,  .949} TGL~\cite{fernandez2024gradient} & \textcolor{red}{\ding{52}} &  &  & CPT & GPT-2 base, GPT-2 large, GPT-Neo & Backbone & - \\ 

       LOIRE~\cite{hanloire} & \textcolor{red}{\ding{52}} & \ding{52} &  & CPT & GPT-2 small, GPT-2 large, BERT-base & Backbone & - \\

       \rowcolor[rgb]{ .949,  .949,  .949} AdapterCL~\cite{madotto2020continual} & \textcolor{red}{\ding{52}} &  &  & CToD & GPT-2 & PEFT & \href{https://github.com/andreamad8/ToDCL}{Link} \\ 

       SAPT~\cite{zhao2024sapt} & \textcolor{red}{\ding{52}} & \ding{52} &  & CIT & T5-large, T5-XL, T5-XXL, LLaMA-2-7B, LLaMA-2-13B & PEFT & \href{https://github.com/circle-hit/SAPT}{Link} \\ 
       \rowcolor[rgb]{ .949,  .949,  .949} MoRAL~\cite{yang2024moral} & \textcolor{red}{\ding{52}} &  &  & CPT & TinyLlama-1.1B, Phi-2-2.7B, LLaMA-2-7B & PEFT & - \\ 
       I-LoRA~\cite{ren2024analyzing} & \textcolor{red}{\ding{52}} &  & \ding{52} & CIT & LLaMA-2-7B & PEFT & \href{https://github.com/which47/LLMCL}{Link} \\ 
       \rowcolor[rgb]{ .949,  .949,  .949} SLIM~\cite{han2024slim} & \textcolor{red}{\ding{52}} &  &  & CIT & OpenChat-8B & PEFT & - \\ 

       SEE~\cite{wang2025see} & \textcolor{red}{\ding{52}} &  & \ding{52} & CIT & LLaMA-2-7B, LLaMA-2-7B-Chat, Alpaca-7B, LLaMA3.1-8B, Qwen2.5-7B, Mistral-7B-v0.3 & PEFT & \href{https://github.com/Linzwcs/SEE}{Link} \\ 
       
       \rowcolor[rgb]{ .949,  .949,  .949} P-Prompts~\cite{razdaibiedina2023progressive} & \textcolor{red}{\ding{52}} &  &  & CIT & BERT-base, T5-large & PEFT & \href{https://github.com/arazd/ProgressivePrompts}{Link} \\ 
       Q-Tuning~\cite{guo2024q} & \textcolor{red}{\ding{52}} & \ding{52} &  & CIT & BERT-base, T5-large & PEFT & - \\ 
       \rowcolor[rgb]{ .949,  .949,  .949} WAVE-CRE~\cite{le2025adaptive} & \textcolor{red}{\ding{52}} &  &  & CRE & BERT-base & PEFT & \href{https://github.com/mrshaw01/AdaptivePromptingCRE}{Link} \\ 

       O-LoRA~\cite{wang2023orthogonal} & \ding{52} & \textcolor{red}{\ding{52}} &  & CIT & T5-large, LLaMA-7B & PEFT & \href{https://github.com/cmnfriend/O-LoRA}{Link} \\ 

       \rowcolor[rgb]{ .949,  .949,  .949} DAS~\cite{ke2023continual} &  & \textcolor{red}{\ding{52}} &  & CPT & RoBERTa-base & Backbone & \href{https://github.com/UIC-Liu-Lab/ContinualLM}{Link} \\ 

       ARPER~\cite{mi2020continual} &  & \textcolor{red}{\ding{52}} & \ding{52} & CToD  & SCLSTM & Backbone & \href{https://github.com/MiFei/Continual-Learning-for-NLG}{Link} \\

       \rowcolor[rgb]{ .949,  .949,  .949} SEEKR~\cite{he2024seekr} &  & \textcolor{red}{\ding{52}} & \ding{52} & CIT & LLaMA-2-7B-Chat, Vicuna-7B-v1.5, Vicuna-13B-v1.5 & Backbone & \href{https://github.com/jinghan1he/SEEKR}{Link} \\

       TaSL~\cite{feng2024tasl} &  & \textcolor{red}{\ding{52}} &  & CToD & T5-small, T5-base, T5-large, LLaMA-7B & Backbone \& PEFT & \href{https://github.com/WoodScene/TaSL}{Link} \\

       \rowcolor[rgb]{ .949,  .949,  .949} DYNAINST~\cite{mok2023large} &  & \textcolor{red}{\ding{52}} & \ding{52} & CIT & BART-base & Backbone & - \\

       DaCL~\cite{wu2023enhancing} &  & \textcolor{red}{\ding{52}} & \ding{52} & CRE & BERT-base & Backbone & \href{https://github.com/CuteyThyme/Noisy-CRE}{Link} \\

       \rowcolor[rgb]{ .949,  .949,  .949} FVG~\cite{jiang2025unlocking} &  & \textcolor{red}{\ding{52}} &  &  CIT & LLaMA-2-7B-Chat, LLaMA-2-13B-Chat, LLaMA-3-8B-Chat, Mistral-7B-instruct-v2 & PEFT & - \\ 

       Velocitune~\cite{luo2024velocitune} &  & \textcolor{red}{\ding{52}} & \ding{52} &  CPT & CodeLLaMA-7B, LLaMA3-8B, Mistral-7B & Backbone & - \\ 

       \rowcolor[rgb]{ .949,  .949,  .949} Recurrent-KIF~\cite{feng2025recurrent} &  & \textcolor{red}{\ding{52}} & \ding{52} &  CIT & T5-large, FLAN-T5-XL, LLaMA-2-7B, LLaMA-2-13B & PEFT & \href{https://github.com/WoodScene/Recurrent_KIF}{Link} \\ 

       CT0~\cite{scialom2022fine} &  &  & \textcolor{red}{\ding{52}} & CIT  & T5-small, T5-3B & Backbone & \href{https://github.com/ThomasScialom/T0_continual_learning}{Link} \\

       \rowcolor[rgb]{ .949,  .949,  .949} Huang et al.~\cite{huang2024towards} &  &  & \textcolor{red}{\ding{52}} & CIT & OPT-125M, OPT-350M, OPT-1.3B, OPT-2.7B, OPT-6.7B, OPT-13B & Backbone & - \\

       Anh et al.~\cite{anh2025mutual} &  & \ding{52} & \textcolor{red}{\ding{52}} & CRE & BERT-base, LLM2Vec & Backbone & - \\

       \rowcolor[rgb]{ .949,  .949,  .949} InsCL~\cite{wang2024inscl} &  &  & \textcolor{red}{\ding{52}} & CIT & LLaMA-7B  & Backbone & \href{https://github.com/OPPO-Mente-Lab/InsCL}{Link} \\

       HESIT~\cite{chen2024overcoming} &  &  & \textcolor{red}{\ding{52}} & CToD & GPT-2 & Backbone & - \\

       \rowcolor[rgb]{ .949,  .949,  .949} Sun et al.~\cite{sun2024reviving} &  &  & \textcolor{red}{\ding{52}} & CIT & Qwen2-0.5B, Qwen2-1.5B, LLaMA-2-7B, LLaMA-2-13B, Mistral-7B & Backbone & \href{https://github.com/DIRECT-BIT/Reviving-Dormant-Memories}{Link} \\

       D-CPT Law~\cite{que2024d} &  &  & \textcolor{red}{\ding{52}} & CPT & Qwen1.5-0.5B, Qwen1.5-1.8B, Qwen1.5-4B & Backbone & - \\

       \rowcolor[rgb]{ .949,  .949,  .949} KPIG~\cite{he2024don} & & \ding{52} & \textcolor{red}{\ding{52}} & CIT & LLaMA-2-7B-Chat, baichuan-vicuna-chinese-7b & Backbone & - \\

       Guo et al.~\cite{guo2024efficient} & & & \textcolor{red}{\ding{52}} & CPT & OpenLLaMA-3B, LLaMA-3-8B & Backbone & - \\

       \rowcolor[rgb]{ .949,  .949,  .949} LAMOL~\cite{sun2019lamol} &  &  & \textcolor{red}{\ding{52}} & CIT &  GPT-2 small  & Backbone & \href{https://github.com/chho33/LAMOL}{Link} \\

       HMI-LAMOL~\cite{maekawa2023generative} &  &  & \textcolor{red}{\ding{52}} & CIT & GPT-2 small, BERT-base & Backbone & \href{https://github.com/arumaekawa/GR-HMI}{Link} \\

       \rowcolor[rgb]{ .949,  .949,  .949} PCLL~\cite{zhao2022prompt} & \ding{52} & \ding{52} & \textcolor{red}{\ding{52}} & CToD & GPT-2 small & PEFT & \href{https://github.com/AlibabaResearch/DAMO-ConvAI/tree/main/pcll}{Link} \\

       DCL~\cite{zeng2023continual} & \ding{52} & \ding{52} & \textcolor{red}{\ding{52}} & CToD & GPT-2 small & PEFT & - \\

       \rowcolor[rgb]{ .949,  .949,  .949} SSR~\cite{huang2024mitigating} & & & \textcolor{red}{\ding{52}} & CIT & LLaMA-2-7B, LLaMA-2-7B-Chat, Alpaca-7B & Backbone & \href{https://github.com/DeepLearnXMU/SSR}{Link} \\

       SynE~\cite{chen2024towards} & & & \textcolor{red}{\ding{52}} & CPT & LLaMA-3-8B, TinyLlama & Backbone & \href{https://github.com/RUC-GSAI/Llama-3-SynE}{Link} \\
       
    \bottomrule
    \end{tabular}}
    \label{tab:llm}
\end{table*}

\subsection{Architecture-based Approach}
\label{LLM:Architecture-based}

\subsubsection{Backbone Expansion}
\label{LLM:model-based}
Backbone expansion methods dynamically adapt network architectures by modifying sparsity, increasing capacity, or adding modular components, enabling the integration of new knowledge without full retraining. As shown in Figure~\ref{fig:llm}~(a), these methods restructure internal components to strike a balance between knowledge retention and adaptation. For instance, TPEM~\cite{geng2021continual} introduces a prune-expand-mask framework for task-oriented dialogue, iteratively pruning redundant weights, retraining sparse subnets, and expanding hidden dimensions based on pruning ratios and task demands.
Zheng et al.~\cite{zheng2025spurious} identify that continual learning degradation often stems from misalignment rather than forgetting, and propose a layer-freezing strategy to stabilize low-level representations, consistently improving performance across CL benchmarks.
TGL~\cite{fernandez2024gradient} leverages gradient probing to localize time-sensitive knowledge in LLMs, updating only salient layers. This targeted adaptation improves efficiency over generic pretraining or editing methods and reduces forgetting.
LOIRE~\cite{hanloire} introduces residual plug-in layers for depth-wise expansion while preserving original functionality. It jointly scales hidden dimensions, FFNs, and attention heads, and employs iterative self-distillation to retain knowledge across distributional shifts.

\subsubsection{LoRA Expansion}
\label{LLM:adapter-based}
To enable efficient knowledge acquisition while preserving the backbone’s capabilities, parameter-efficient fine-tuning (PEFT) methods, exemplified by LoRA, have been widely adopted. These methods inject lightweight, task-specific modules—typically low-rank weight updates—allowing new task adaptation with minimal interference to prior knowledge.
AdapterCL~\cite{madotto2020continual} incorporates residual adapters to disentangle task-specific knowledge and employs a perplexity-based strategy for adapter selection during inference.
SAPT~\cite{zhao2024sapt} aligns PET block learning and selection through a shared attentive module, using generated pseudo-samples to retrieve cross-task shared attention, thus enabling task-free robust adaptation.
MoRAL~\cite{yang2024moral} combines LoRA with mixture-of-experts routing, achieving dynamic expert selection with efficient adaptation.
I-LoRA~\cite{ren2024analyzing} proposes a dual-memory replay framework comprising a fast learner for rapid adaptation and a slow learner for long-term retention.
SLIM~\cite{han2024slim} introduces input-dependent LoRA activation via a learnable router and clustering mechanism, enabling selective routing between identity paths and LoRA modules based on input domain characteristics.
SEE~\cite{wang2025see} introduces sequential expert routing for continual instruction tuning, where each expert autonomously decides query handling via distributed routing. Unlike MoE methods requiring retrained routers, SEE trains only new experts per task while maintaining base model generalization.

\subsubsection{Prompt Expansion}
\label{LLM:prompt-based}
As an alternative PEFT strategy for LLMs, prompt-based methods modularize task knowledge into prompts~\cite{lester2021power}, enabling lightweight adaptation, selective updates, and reduced forgetting during continual learning.
P-Prompts~\cite{razdaibiedina2023progressive} allocate a unique prompt for each task and concatenate them in sequence before inputting them into a frozen backbone. By freezing task-specific prompts post-training, it avoids forgetting while promoting forward transfer through prompt reuse.
Q-Tuning~\cite{guo2024q} introduces a continual prompt tuning framework with a prompt queue that retains previous tasks. It adds new prompts for incoming tasks, applies low-rank adaptive aggregation to reweight relevant past knowledge, uses a PCA-based eviction strategy to control queue size, and includes a globally shared prefix prompt with memory retention regularization to maintain stability.
WAVE-CRE~\cite{le2025adaptive} targets continual relation extraction by combining task-specific prompt pools with generative knowledge consolidation. Each task maintains its own prompt set to capture intra-task diversity, while a generative model integrates prior knowledge into shared parameters, mitigating forgetting without storing previous data.

\subsection{Regularization-based Approach}
\label{LLM:Regularization-based}
Regularization-based approaches, as shown in Figure~\ref{fig:llm}~(b), alleviate catastrophic forgetting by imposing constraints during training to preserve important parameters or representations from previous tasks. For example, O-LoRA~\cite{wang2023orthogonal} adopts task-specific LoRA modules with orthogonal regularization, enforcing orthogonality across low-rank updates to reduce interference between tasks and preserve prior knowledge.
DAS~\cite{ke2023continual} proposes a domain-adaptive pretraining framework with soft-masking to control parameter updates based on unit importance. It retains general knowledge using robustness-based metrics and enables cross-domain transfer without domain labels via contrastive knowledge comparison.
ARPER~\cite{mi2020continual} builds on elastic weight consolidation~\cite{kirkpatrick2017overcoming} by assigning dynamic penalties to parameters based on vocabulary shifts, focusing protection on knowledge-critical weights across tasks.
SEEKR~\cite{he2024seekr} preserves key attention head outputs via knowledge distillation and introduces task sensitivity and forgettability scores to prioritize heads most valuable for long-term retention.
TaSL~\cite{feng2024tasl} scores parameter importance group-wise using gradient signals, then applies selective model averaging to retain shared and task-specific knowledge, improving both forgetting mitigation and bidirectional transfer.
DYNAINST~\cite{mok2023large} combines local minima regularization with dynamic instruction replay. It introduces a KL-divergence loss to smooth model output distributions, enhancing generalization and robustness in lifelong in-context learning.
NaCL~\cite{wu2023enhancing} presents a contrastive continual learning method for relation extraction, using an auxiliary model to identify clean samples and adversarially aligning noisy data in feature space, thereby boosting representation robustness.
FVG~\cite{jiang2025unlocking} introduces function vector-guided training to maintain activation consistency and KL alignment across tasks, identifying biased function activations rather than parameter drift as the main cause of forgetting.
Velocitune~\cite{luo2024velocitune} dynamically adjusts domain weights during continual pre-training by measuring and balancing learning velocities across domains. It uses scaling laws to predict target losses and prioritizes slower-learning domains through velocity-guided updates.
Recurrent-KIF~\cite{feng2025recurrent} dynamically estimates parameter importance through iterative inner-outer loops, enabling adaptive knowledge fusion via importance-based binary masking while mitigating catastrophic forgetting and enhancing knowledge transfer across sequential tasks.

\subsection{Replay-based Approach}
\label{LLM:Replay-based}
\subsubsection{Direct Replay}
\label{LLM:exp_reg}
Replay-based approaches, illustrated in Figure~\ref{fig:llm}~(c), alleviate catastrophic forgetting by storing and reusing selected samples or their compressed representations from prior tasks during the training of new ones.
CT0~\cite{scialom2022fine} introduces a lightweight rehearsal mechanism by replaying 1\% of historical instruction data per task. Despite its simplicity, this approach enables the model to maintain instruction-following capabilities across tasks with minimal memory overhead.
Huang et al.~\cite{huang2024towards} enhance replay with a tool-augmented mechanism and a compact buffer, improving the model’s ability to generalize to novel tasks while mitigating forgetting through task-complementary tools.
Anh et al.~\cite{anh2025mutual} construct dual-label inputs combining old and new data, enhance robustness via adversarial perturbations, and apply task-aware loss functions alongside Sharpness-Aware Minimization for improved generalization.
Several methods further adopt dynamic replay strategies.
InsCL~\cite{wang2024inscl} adjusts the replay ratio in real-time based on task similarity and sample-level informativeness. It introduces an instruction informativeness score to filter and favor diverse, semantically rich exemplars.
HESIT~\cite{chen2024overcoming} selects replay samples by analyzing their gradient impact early in training, ensuring that only the most influential instances are retained—maximizing memory utility while minimizing redundancy.
Sun et al.~\cite{sun2024reviving} identify reasoning degradation from ineffective prompts and propose a Rationale-Guidance Difficulty metric to guide selective replay of reasoning-critical data.
Other works focus on optimizing replay ratios between past and present data.
D-CPT Law~\cite{que2024d} extends the Chinchilla scaling law\cite{hoffmann2022training} by introducing a mixture ratio parameter that balances domain-specific and general corpora. Through a Domain-specific Learnable Coefficient, it dynamically tunes the data composition to improve cross-domain generalization under constrained training resources.
KPIG~\cite{he2024don} improves continual instruction tuning by dynamically replaying low-information-gain tasks and refining objectives with key-part masking and JSD-based regularization, enhancing both instruction-following and generalization.
Guo et al.~\cite{guo2024efficient} mitigate the stability gap in continual pre-training via multi-epoch training on high-quality subsets, dynamic data replay guided by information gain, and preservation of pre-training data distributions to balance plasticity and stability while boosting domain performance.

\subsubsection{Generative Replay}
\label{LLM:gen_reg}
Generative replay methods use LLMs to synthesize pseudo-samples from previous tasks, enabling knowledge retention without storing original data.
LAMOL~\cite{sun2019lamol} introduces a generation token to guide pseudo-sample creation during new task learning, maintaining prior knowledge with minimal overhead.
HMI-LAMOL~\cite{maekawa2023generative} enhances this by encoding training data into compressed “memory traces” using BERT, which conditions generation and mitigates sample imbalance and quality degradation.
PCLL~\cite{zhao2022prompt} constructs natural language prompts and uses a conditional variational autoencoder to generate task-specific pseudo-data from latent representations.
DCL~\cite{zeng2023continual} replaces the Gaussian prior with a Dirichlet distribution to better capture task differences and introduces Jensen–Shannon distillation to improve knowledge preservation.
SSR~\cite{huang2024mitigating} leverages in-context learning to self-generate synthetic samples, refine them with the latest model, and select high-quality data via clustering—achieving replay without external generators or stored data.
SynE~\cite{chen2024towards} boosts LLaMA-3’s Chinese proficiency and scientific reasoning by employing synthetic QA generation, topic-driven data mixing, perplexity-based curriculum learning, and dynamic bilingual adaptation, while mitigating catastrophic forgetting through targeted knowledge retention strategies.

\subsection{Discussions}

This section reviewed the application of continual learning in large language models, focusing on three main approaches: architecture-based, regularization-based, and replay-based methods. Among them, architectural expansion and replay have emerged as dominant strategies due to the task diversity and semantic complexity inherent to LLMs. The former enhances task specialization by introducing lightweight, modular components such as plug-in LoRA adapters, offering scalability and modularity. Replay-based methods, on the other hand, leverage the strong generalization capabilities of LLMs, requiring only a small number of previous samples or features to effectively mitigate forgetting. While regularization has proven effective in traditional networks, its adoption in LLMs remains limited due to challenges in tuning and stability. Nonetheless, the integration of regularization with architectural expansion or replay strategies, such as applying knowledge-preserving constraints during multitask pretraining or dynamically evaluating parameter importance during updates, holds great potential for developing more robust and generalizable continual learning frameworks in LLMs.

\section{Continual Learning for Multimodal Large Language Models}
\label{Sec:MLLM}

Recent Multimodal Large Language Models~\cite{zeng2025parameter}, built upon advanced Large Language Models, have demonstrated impressive visual understanding capabilities by integrating visual encoders, cross-modal projectors, and leveraging a large-scale image-text corpus during training. Continual learning in MLLMs seeks to equip them with task-specific abilities across domains such as healthcare~\cite{saab2024capabilities, chen2024huatuogpt}, finance~\cite{bhatia2024fintral}, and autonomous driving~\cite{xing2025openemma}, without forgetting their previously learned knowledge. In contrast to discriminative models that predict output through classifiers, MLLMs generate responses by next-token prediction in an autoregressive way, making most continual learning methods designed for discriminative models less applicable. Additionally, compared to LLMs, MLLMs need to simultaneously process information from multiple modalities~\cite{zeng2024clip}, thus posing a greater challenge.

In this section, we begin by introducing the setup for an MLLM continual learning task~(Section \ref{MLLM:setup}), which includes training objectives, common model backbones, and existing benchmarks. Subsequently, the existing literature is categorized into three strategies: architecture-based~(Section \ref{MLLM:art}), regularization-based~(Section \ref{MLLM:reg}), and replay-based~(Section \ref{MLLM:replay}) approaches.

\subsection{Problem Setup}
\label{MLLM:setup}
\textbf{Training objectives.} 
The visual instruction tuning dataset $\mathcal{D} = \{ (\textbf{X}_{v}^{j}, \textbf{X}_{ins}^{j}, \textbf{X}_{a}^{j})_{j=1}^{N}\}$ used for fine-tuning a MLLM $f_{\boldsymbol{\theta}}$ typically consists of image input $\textbf{X}_{v}$, user instruction $\textbf{X}_{ins}$ and target answer $\textbf{X}_{a}$, where $N$ represents the total number of training samples. Given an image-instruction pair with a response of length $L$, the MLLM is trained to predict each token in the answer autoregressively, conditioned on all preceding tokens:
\begin{equation}
        p(\textbf{X}_{a}|\textbf{X}_v, \textbf{X}_{ins}) = \prod_{i=1}^{L}p_{\boldsymbol{\theta}}(x_i|\textbf{X}_v, \textbf{X}_{ins},\textbf{X}_{a, <i}),
\label{eq:lmm}
\end{equation}
\noindent
where $\theta$ denotes the trainable parameters during fine-tuning, $\textbf{X}_{a,<i}$ are the response tokens from all previous turns before the current prediction token $x_i$. In the continual learning framework for MLLMs, the optimization objective and loss function maintain the same formulation as the LLM case in \eq~(\ref{eq:llm-loss}), with the distinction that the input-output pairs now involve multimodal data. Here, $\boldsymbol{x}$ and $\boldsymbol{y}$ represent the multimodal inputs $(\textbf{X}_v, \textbf{X}_{ins})$ and the target answer $\textbf{X}_{a}$, respectively.

\noindent
\textbf{Backbones.} To process inputs from multiple modalities, MLLMs typically first extract modality-specific features using dedicated encoders. These features are then concatenated and passed into a pre-trained language model that generates responses autoregressively. Based on current literature, we summarize the MLLM architectures commonly adopted in continual learning as follows:
\begin{itemize}
    \item \textbf{VL-T5}~\cite{zhang2023vqacl} uses two separate visual encoders: Faster R-CNN for region features in visual question answering, and 3D ResNeXt-101 for capturing video motion in video question answering. Its language model consists of 12 encoder-decoder blocks with 12-head multi-head attention layers and an embedding dimension of 768.
    \item \textbf{InstructBLIP}~\cite{dai2023instructblip} leverages a Query Transformer (Q-Former) to extract visual features from a frozen ViT-G/14 image encoder. The Q-Former uses $K$ learnable query embeddings that interact with image features through cross-attention. The resulting outputs are linearly projected and then passed into a frozen LLM, typically FlanT5-XL. During training, only the projection layer of the Q-Former and the task-specific query keys are updated.
    \item \textbf{LLaVA}~\cite{liu2023visual} series is known for its strong visual understanding and streamlined design. It uses Vicuna as the LLM and a pre-trained CLIP ViT-L/14@336 visual encoder. Visual features are aligned with text through a simple linear projection into a shared representation space.
\end{itemize}

In addition to the MLLM backbones mentioned above, some methods incorporate customized architectural designs tailored to their specific settings (e.g., Qwen-VL series~\cite{Qwen-VL, bai2025qwen2}, MiniGPT-v2~\cite{chen2023minigpt}, InternVL~\cite{chen2024internvl}, etc.), which are detailed in Table~\ref{tab:mllm}.

\begin{figure*}[t]
    \centering
    \includegraphics[width=0.98\linewidth]{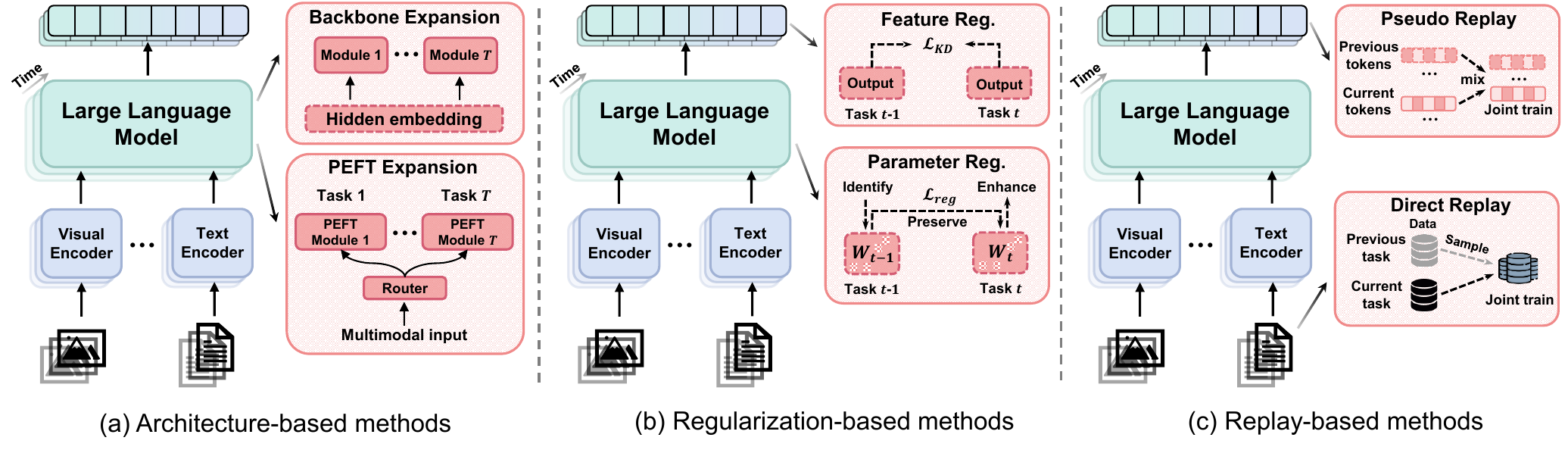}
    \caption{Three types of continual learning approaches in Multimodal Large Language Models.}
    \vspace{-10pt}
    \label{fig:mllm}
\end{figure*}

\noindent
\textbf{Benchmarks.} 
To assess the continual learning capabilities of MLLMs, several representative benchmarks have been developed. Below, we highlight key benchmarks; for detailed datasets and usage, we refer readers to
 their original papers.
\begin{itemize}
    \item \textbf{CLiMB}~\cite{srinivasan2022climb} evaluates continual learning in vision-and-language tasks through two phases: (1) Upstream Learning, where a pre-trained multimodal model is trained sequentially on tasks to assess forgetting and knowledge transfer; and (2) Downstream Transfer, which measures low-shot adaptation to both multimodal and unimodal tasks. Results indicate that existing methods struggle to generalize across task sequences and adapt effectively in low-shot scenarios.
    \item \textbf{VQACL}~\cite{zhang2023vqacl} constructs a dual-level task sequence from VQA2.0 and NExT-QA, with outer linguistic-driven tasks representing reasoning skills (e.g., counting) and inner visual-driven subtasks focused on specific object groups. VQACL tests compositionality by withholding one visual subtask per linguistic task during training and using it for evaluation.
    \item \textbf{CoIN}~\cite{chen2024coin} includes 8 curated datasets covering general image QA, visual reasoning, knowledge-grounded QA, and so on. It evaluates MLLMs from two perspectives: (1) Truth Alignment, which measures adherence to correct outputs and instructions; and (2) Reasoning Capability, which assesses the ability to retain and integrate knowledge over time.
    \item \textbf{MLLM-CL}~\cite{hongbozhao25} introduces two settings: domain continual learning (DCL) focusing on mainstream domains (remote sensing, medical, autonomous driving, science, and finance) with IID evaluations, and ability continual learning (ACL) targeting fundamental abilities (OCR, math, visual perception, GUI agents) in non-IID scenarios, providing a comprehensive, realistic continual learning evaluation.
\end{itemize}

\subsection{Architecture-based Approach}
\label{MLLM:art}

\subsubsection{Backbone Expansion}
\label{MLLM:backbone_exp}

As shown in Figure~\ref{fig:mllm}~(a), backbone expansion methods dynamically extend model architectures through modular components, enabling adaptive learning across diverse tasks. Eproj~\cite{he2023continual} establishes the first continual instruction tuning benchmark for MLLMs, addressing catastrophic forgetting via dynamic Q-Former layer selection based on multimodal similarity scoring, combined with adaptive regularization and strategic old-sample replay. Meanwhile, TAM-CL~\cite{cai2023task} introduces a transformer-based continual learning framework that expands capacity through task-specific attention layers with learnable tokens, while preserving knowledge via teacher-student distillation and selective experience replay from a compact memory buffer.

\begin{table*}[!t]
    \centering
    \caption{Summary of continual learning methods for Multimodal Large Language Models. Taxonomy: \textcolor{red}{\ding{52}} indicates the primary solution, and \ding{52} indicates the supporting method. Benchmark: Composite benchmark refer to method that combine multiple datasets without a unified name. \textbf{Tag}: \encircle[fill=lightcoral, text=white]{I} = \underline{I}mage, \encircle[fill=brightlavender, text=white]{T} = \underline{T}ext, \encircle[fill=lightgreen, text=white]{V} = \underline{V}ideo, \encircle[fill=capri, text=white]{A} = \underline{A}udio.}
    \resizebox{0.98\textwidth}{!}{
    \begin{tabular}{lcccllccc}
    \toprule
       \multirow{2}{*}{\textbf{Methods}} & \multicolumn{3}{c}{\textbf{Taxonomy}} & \multirow{2}{*}{\textbf{Benchmark}} & \multirow{2}{*}{\textbf{Backbone / Basic Models}} & \multirow{2}{*}{\textbf{Modality}} & \multirow{2}{*}{\textbf{Implement}} & \multirow{2}{*}{\textbf{Code}} \\ 
       \cmidrule{2-4}
       & \textbf{Arch.} & \textbf{Reg.} & \textbf{Rep.}  \\ \midrule

       \rowcolor[rgb]{ .949,  .949,  .949} Eproj~\cite{he2023continual} & \textcolor{red}{\ding{52}} & \ding{52} & \ding{52} & Composite & InstrutBLIP, BILP2 & \encircle[fill=lightcoral, text=white]{I} \encircle[fill=brightlavender, text=white]{T} & Backbone & - \\

       TAM-CL~\cite{cai2023task} & \textcolor{red}{\ding{52}} & \ding{52} & \ding{52} & Composite & ViLT & \encircle[fill=lightcoral, text=white]{I} \encircle[fill=brightlavender, text=white]{T} & Backbone & \href{https://github.com/YuliangCai2022/TAMCL.git.}{Link} \\
       
       \rowcolor[rgb]{ .949,  .949,  .949} MoELoRA~\cite{chen2024coin} & \textcolor{red}{\ding{52}} & & & CoIN & LLaVA-1.5-7B, LLaVA-1.5-13B, Qwen-VL-Chat, MiniGPT-v2 & \encircle[fill=lightcoral, text=white]{I} \encircle[fill=brightlavender, text=white]{T} & PEFT & \href{https://github.com/zackschen/CoIN}{Link} \\

       Continual LLaVA~\cite{cao2024continual} & \textcolor{red}{\ding{52}} & & & COAST & LLaVA-1.5-7B & \encircle[fill=lightcoral, text=white]{I} \encircle[fill=brightlavender, text=white]{T} & PEFT & \href{https://github.com/mengcaopku/Continual-LLaVA}{Link} \\

       \rowcolor[rgb]{ .949,  .949,  .949} HiDe-LLaVA~\cite{guo2025hide} & \textcolor{red}{\ding{52}} & & & CoIN, UCIT & LLaVA-1.5-7B, InternVL-Chat-7B & \encircle[fill=lightcoral, text=white]{I} \encircle[fill=brightlavender, text=white]{T} & PEFT & \href{https://github.com/Ghy0501/HiDe-LLaVA}{Link} \\

       LLaCA~\cite{qiao2024llaca} & \textcolor{red}{\ding{52}} & & & CoIN & LLaVA-1.5-7B, LLaVA-1.5-13B, Qwen-VL-Chat & \encircle[fill=lightcoral, text=white]{I} \encircle[fill=brightlavender, text=white]{T} & PEFT & \href{https://jingyangqiao.github.io/}{Link} \\

       \rowcolor[rgb]{ .949,  .949,  .949} SMoLoRA~\cite{wang2024separable} & \textcolor{red}{\ding{52}} & & & CVIT & LLaVA-1.5-7B & \encircle[fill=lightcoral, text=white]{I} \encircle[fill=brightlavender, text=white]{T} & PEFT & - \\

       Cl-MoE~\cite{huai2025cl} & \textcolor{red}{\ding{52}} & & & VQACL & VL-T5, LLaVA-1.5-7B & \encircle[fill=lightcoral, text=white]{I} \encircle[fill=brightlavender, text=white]{T} & PEFT & \href{https://github.com/ECNU-ICALK/CL-MoE}{Link} \\

       \rowcolor[rgb]{ .949,  .949,  .949} LLaVA-CMoE~\cite{zhao2025llava} & \textcolor{red}{\ding{52}} & & \ding{52} & CoIN & LLaVA-1.5-7B & \encircle[fill=lightcoral, text=white]{I} \encircle[fill=brightlavender, text=white]{T} & PEFT & - \\

       DISCO~\cite{guo2025federated} & \textcolor{red}{\ding{52}} & & & FCIT & LLaVA-1.5-7B & \encircle[fill=lightcoral, text=white]{I} \encircle[fill=brightlavender, text=white]{T} & PEFT & \href{https://github.com/Ghy0501/FCIT}{Link} \\

       \rowcolor[rgb]{ .949,  .949,  .949} MR-LoRA~\cite{hongbozhao25} &\textcolor{red}{\ding{52}}  &  & \ding{52}  & MLLM-CL & LLaVA-1.5-7B & \encircle[fill=lightcoral, text=white]{I} \encircle[fill=brightlavender, text=white]{T} & PEFT & \href{https://github.com/bjzhb666/MLLM-CL}{Link} \\

       D-MoLE~\cite{ge2025dynamic} &\textcolor{red}{\ding{52}} &  &  & CMIT & InternVL2-2B & \encircle[fill=lightcoral, text=white]{I} \encircle[fill=brightlavender, text=white]{T} & PEFT & \href{https://github.com/gcd19/D-MoLE}{Link} \\

       \rowcolor[rgb]{ .949,  .949,  .949} BranchLoRA~\cite{zhang2025enhancing} &\textcolor{red}{\ding{52}} &  &  & CoIN & LLaVA-1.5-7B, LLaVA-1.5-13B & \encircle[fill=lightcoral, text=white]{I} \encircle[fill=brightlavender, text=white]{T} & PEFT & \href{https://github.com/BladeDancer957/BranchLoRA}{Link} \\

       ProgLoRA~\cite{yu2025progressive} &\textcolor{red}{\ding{52}} &  &  & CoIN & LLaVA-1.5-7B & \encircle[fill=lightcoral, text=white]{I} \encircle[fill=brightlavender, text=white]{T} & PEFT & \href{https://github.com/ku-nlp/ProgLoRA}{Link} \\

       \rowcolor[rgb]{ .949,  .949,  .949} TRIPLET~\cite{qian2023decouple} & \textcolor{red}{\ding{52}} & \ding{52} & & Composite & ALBEF, FLAVA & \encircle[fill=lightcoral, text=white]{I} \encircle[fill=brightlavender, text=white]{T} & PEFT & - \\

       ModalPrompt~\cite{zeng2024modalprompt} & \textcolor{red}{\ding{52}} & & & CoIN & LLaVA-1.5-7B & \encircle[fill=lightcoral, text=white]{I} \encircle[fill=brightlavender, text=white]{T} & PEFT & \href{https://github.com/AuroraZengfh/ModalPrompt}{Link} \\

       \rowcolor[rgb]{ .949,  .949,  .949} Fwd-Prompt~\cite{zheng2024beyond} & \textcolor{red}{\ding{52}} & & & Composite & InstructBLIP, BLIP2 & \encircle[fill=lightcoral, text=white]{I} \encircle[fill=brightlavender, text=white]{T} & PEFT & - \\

       ColPro~\cite{cai2024empowering} & \textcolor{red}{\ding{52}} & \ding{52} & & Composite & LLaMA-7B, CLIP ViT-L/14 & \encircle[fill=lightgreen, text=white]{V} \encircle[fill=brightlavender, text=white]{T} & PEFT & \href{https://github.com/caicch/ColPro}{Link} \\

       \rowcolor[rgb]{ .949,  .949,  .949} CluMo~\cite{cai2024clumo} & \textcolor{red}{\ding{52}} & \ding{52} & & CLOVE & ALBEF & \encircle[fill=lightcoral, text=white]{I} \encircle[fill=brightlavender, text=white]{T} & PEFT & \href{https://github.com/YuliangCai2022/CLUMO}{Link} \\

       \rowcolor[rgb]{ .949,  .949,  .949}
       SCD~\cite{lao2023multi} &  & \textcolor{red}{\ding{52}} & & MDL-VQA & ViLT & \encircle[fill=lightcoral, text=white]{I} \encircle[fill=brightlavender, text=white]{T} & Backbone & - \\

       CS-VQLA~\cite{bai2023revisiting} &  & \textcolor{red}{\ding{52}} & & Composite & ResNet, VisualBERT & \encircle[fill=lightcoral, text=white]{I} \encircle[fill=brightlavender, text=white]{T} & Backbone & \href{https://github.com/longbai1006/CS-VQLA}{Link} \\

       \rowcolor[rgb]{ .949,  .949,  .949}
       Chen et al~\cite{chen2024llm} &  & \textcolor{red}{\ding{52}} & & Composite & InstructBLIP & \encircle[fill=lightcoral, text=white]{I} \encircle[fill=brightlavender, text=white]{T} & Backbone & - \\

       CrossSDC~\cite{pian2024continual} &  & \textcolor{red}{\ding{52}} & & Composite & iQuery & \encircle[fill=lightgreen, text=white]{V} \encircle[fill=capri, text=white]{A} & Backbone & \href{https://github.com/weiguoPian/ContAV-Sep_NeurIPS2024}{Link} \\

       \rowcolor[rgb]{ .949,  .949,  .949}
       MAFED~\cite{nikandrou2024enhancing} &  & \textcolor{red}{\ding{52}} & \ding{52} & Composite & UNITER, ViLT, VL-Pythia & \encircle[fill=lightcoral, text=white]{I} \encircle[fill=brightlavender, text=white]{T} & Backbone & \href{https://github.com/MalvinaNikandrou/mafed}{Link} \\

       MoInCL~\cite{pian2024modality} &  & \textcolor{red}{\ding{52}} & \ding{52} & Composite & LLaMA-3.2-1B-Instruct, EVA-CLIP-ViT-G/14, BEAT$_{s_{\text{iter3+}}}$ & \encircle[fill=lightcoral, text=white]{I} \encircle[fill=lightgreen, text=white]{V} \encircle[fill=capri, text=white]{A} \encircle[fill=brightlavender, text=white]{T} & PEFT & - \\

       \rowcolor[rgb]{ .949,  .949,  .949}
       QUAD~\cite{marouf2025no} &  & \textcolor{red}{\ding{52}} & \ding{52} & VQACL & VL-T5 & \encircle[fill=lightcoral, text=white]{I} \encircle[fill=brightlavender, text=white]{T} & Backbone & \href{https://github.com/IemProg/QUAD}{Link} \\

       Model Tailor~\cite{zhu2024model} &  & \textcolor{red}{\ding{52}} &  & Composite & InstructBLIP, LLaVA-1.5-7B & \encircle[fill=lightcoral, text=white]{I} \encircle[fill=brightlavender, text=white]{T} & Backbone \& PEFT & \href{https://github.com/didizhu-zju/Model-Tailor}{Link} \\

       \rowcolor[rgb]{ .949,  .949,  .949}
       SPIDER~\cite{huang2024learn} &  & \textcolor{red}{\ding{52}} &  & Composite & VILA-1.5-3B, LLaVA-1.5-7B & \encircle[fill=lightcoral, text=white]{I} \encircle[fill=brightlavender, text=white]{T} & Backbone & - \\

       SEFE~\cite{chen2025sefe} &  & \textcolor{red}{\ding{52}} &  & CoIN-ASD & LLaVA-1.5-7B & \encircle[fill=lightcoral, text=white]{I} \encircle[fill=brightlavender, text=white]{T} & PEFT & \href{https://github.com/jinpeng0528/SEFE}{Link} \\

       \rowcolor[rgb]{ .949,  .949,  .949} LoRASculpt~\cite{liang2025lorasculpt} &  & \textcolor{red}{\ding{52}} &  & Composite & LLaVA-1.5-7B & \encircle[fill=lightcoral, text=white]{I} \encircle[fill=brightlavender, text=white]{T} & PEFT & \href{https://github.com/LiangJian24/LoRASculpt}{Link} \\

       LLaVA-c~\cite{liu2025llava} &  & \textcolor{red}{\ding{52}} &  & Composite & LLaVA-1.5-7B & \encircle[fill=lightcoral, text=white]{I} \encircle[fill=brightlavender, text=white]{T} & PEFT & - \\

       \rowcolor[rgb]{ .949,  .949,  .949}
       VQACL~\cite{zhang2023vqacl} &  &  & \textcolor{red}{\ding{52}} & VQACL & VL-T5 & \encircle[fill=lightcoral, text=white]{I} \encircle[fill=brightlavender, text=white]{T} & Backbone & \href{https://github.com/zhangxi1997/VQACL}{Link} \\

       SGP~\cite{lei2023symbolic} &  &  & \textcolor{red}{\ding{52}} & CLOVE & UniVQA & \encircle[fill=lightcoral, text=white]{I} \encircle[fill=brightlavender, text=white]{T} & Backbone & \href{https://github.com/showlab/CLVQA}{Link} \\

       \rowcolor[rgb]{ .949,  .949,  .949}
       ProtoGroup~\cite{zhang2025multi} &  &  & \textcolor{red}{\ding{52}} & VQACL & VL-T5 & \encircle[fill=lightcoral, text=white]{I} \encircle[fill=brightlavender, text=white]{T} & Backbone & - \\

       Lin et al~\cite{lin2025vlm} &  & \ding{52} & \textcolor{red}{\ding{52}} & Composite & EM-VLM4AD & \encircle[fill=lightcoral, text=white]{I} \encircle[fill=brightlavender, text=white]{T} & Backbone & - \\

       \rowcolor[rgb]{ .949,  .949,  .949}
       Adapt-$\infty$~\cite{maharana2024adapt} &  &  & \textcolor{red}{\ding{52}} & Composite & LLaVA-1.5-7B & \encircle[fill=lightcoral, text=white]{I} \encircle[fill=brightlavender, text=white]{T} & PEFT & \href{https://github.com/adymaharana/adapt-inf}{Link} \\

       OASIS~\cite{lee2025oasis} &  & \ding{52} & \textcolor{red}{\ding{52}} & MICVIT & LLaVA-1.5-7B, Qwen-VL-2.5-7B & \encircle[fill=lightcoral, text=white]{I} \encircle[fill=brightlavender, text=white]{T} & PEFT & - \\
       
    \bottomrule
    \end{tabular}}
    \vspace{-5pt}
    \label{tab:mllm}
\end{table*}

\subsubsection{LoRA Expansion}
\label{MLLM:lora_exp}
LoRA expansion methods enhance MLLMs by adding lightweight low-rank adapters to a fixed backbone, balancing learning flexibility and memory stability for efficient continual learning.
MoELoRA~\cite{chen2024coin} embeds LoRA modules as experts within a mixture of experts, using a learnable router for dynamic expert selection, evaluated on the CoIN benchmark.
Continual LLaVA~\cite{cao2024continual} introduces COAST, a benchmark assessing domain shifts, capability growth, and dataset evolution, mitigating forgetting via a dual-embedding combining instruction-aware and cross-task contextual features alongside selective LoRA modules.
HiDe-LLaVA~\cite{guo2025hide} decomposes the model via layer-wise CKA similarity into task-specific expansions and general fusion, guided by dual-modality prototype matching for adaptive LoRA selection, with parameter-efficient merging to preserve shared knowledge, supported by the UCIT benchmark.
MR-LoRA~\cite{hongbozhao25} employs architectural decoupling to preserve plasticity and uses multimodal routing based on cross-modal representations for dynamic expert selection in complex scenarios.
LLaCA~\cite{qiao2024llaca} balances stability and plasticity with gradient-guided EMA weight adaptation and Taylor expansion, enhanced by semantic instruction grouping for robust anti-forgetting.
SMoLoRA~\cite{wang2024separable} introduces a separable mixture of low-rank adapters with distinct routing for vision and language tasks, dynamically fusing LoRA blocks to reduce dual forgetting, evaluated on the CVIT benchmark.
CL-MoE~\cite{huai2025cl} proposes dual-router MoE combining instance- and task-level routing, plus momentum-based parameter updates for shared and task-specific experts to balance retention and acquisition.
LLaVA-CMoE~\cite{zhao2025llava} dynamically expands experts based on activation frequency, leveraging hierarchical routing and VAE-modeled task distributions to avoid interference and preserve zero-shot capabilities.
FCIT~\cite{guo2025federated} presents DISCO for federated continual instruction tuning, dynamically managing task subspaces via identity token matching and selective activation, establishing the first FCIT benchmark addressing data heterogeneity and continual learning challenges.
MR-LoRA~\cite{hongbozhao25} combines domain-specific LoRA modules with an MLLM-based routing mechanism to dynamically select optimal experts, preventing catastrophic forgetting via parameter isolation and few-shot adaptation.
D-MoLE~\cite{ge2025dynamic} introduces dynamic layer-wise LoRA expert allocation and gradient-based inter-modal curriculum to optimize continual multimodal instruction tuning, achieving adaptive parameter distribution across transformer layers while balancing modality-specific updates through task-aware difficulty assessment.
BranchLoRA~\cite{zhang2025enhancing} uses a shared matrix $A$ in LoRA modules to capture task-invariant patterns, alongside task-specific matrices $B$ equipped with flexible tuning–freezing mechanisms and task-specific routers, thereby effectively mitigating catastrophic forgetting.
ProgLoRA~\cite{yu2025progressive} employs a progressive LoRA pool with task-aware allocation and recall to mitigate forgetting and enhance knowledge transfer in multimodal continual instruction tuning.

\subsubsection{Prompt Expansion}
\label{MLLM:prompt_exp}
Prompt expansions are a standard technique used to improve the continual learning capability of MLLMs. For example,
TRIPLET~\cite{qian2023decouple} proposes the first multimodal prompt learning framework for continual visual question answering, introducing decoupled prompts across modalities, transformer layers, and tasks, combined with query-matching and cross-modal interactions to model complex dependencies and mitigate forgetting without rehearsal buffers.
ModalPrompt~\cite{zeng2024modalprompt} presents a dual-modality guided prompt learning approach that constructs task-specific prototype prompts to preserve prior knowledge without replay, using image-text guided prompt selection to dynamically match relevant prompts at inference.
Fwd-Prompt~\cite{zheng2024beyond} applies prompt tuning with gradient projection, allocating orthogonal subspaces to prevent interference by projecting gradients to residual spaces for old knowledge retention while reusing core pre-trained knowledge to boost forward transfer.
ColPro~\cite{cai2024empowering} introduces collaborative prompting by integrating task-specific question constraints, knowledge acquisition, and visual temporal awareness prompts into a pre-trained LLM, leveraging complementary learning with General and Expert prompts, negative guiding, autoregressive temporal dynamics, and multimodal distillation to enhance video QA and reduce forgetting.
CluMo~\cite{cai2024clumo} proposes a cluster-based modality fusion prompting method with task-specific visual and textual prompt keys learned via K-means clustering, selecting optimal prompts from a shared pool using a two-stage training and knowledge distillation strategy to improve generalization and mitigate forgetting while keeping the backbone frozen.

\subsection{Regularization-based Approach}
\label{MLLM:reg}

\subsubsection{Feature-Level Regularization}
\label{MLLM:fea_reg}
To overcome catastrophic forgetting, the idea of Knowledge Distillation has also demonstrated strong potential in continual learning approaches for MLLM. One line of work uses current task data directly for distillation, aiming to retain previous knowledge without relying on past data. For example, SCD~\cite{lao2023multi} distills instance- and domain-relevant knowledge by comparing teacher responses to original and counterfactual samples, employing metric learning to avoid harmful domain-specific features while acquiring transferable knowledge. CS-VQLA~\cite{bai2023revisiting} introduces rigidity-plasticity-aware and self-calibrated heterogeneous distillation to balance retention and adaptability in robotic surgery tasks. Chen et al.~\cite{chen2024llm} propose an LLM-assisted distillation loss that leverages LLMs’ generalization to handle domain shifts and data imbalance. CrossSDC~\cite{pian2024continual} preserves cross-modal semantic similarity over time and can be integrated into various frameworks to prevent forgetting in continual audio-visual sound separation.

Another approach uses distillation with past task data or generated pseudo-samples. MAFED~\cite{nikandrou2024enhancing} mixes a small portion of old samples into new task data and applies modality-aware feature distillation, weighting questions and visual tokens by importance. MoInCL~\cite{pian2024modality} proposes a
pseudo-target generation module to synthesize input-target pairs
for various task types. These pseudo samples are then mixed with
current task data for knowledge distillation, effectively mitigating
catastrophic forgetting. QUAD~\cite{marouf2025no} uses only past task questions for regularization to avoid overfitting and introduces attention consistency distillation to preserve intra- and inter-modal attention across tasks.

\subsubsection{Parameter-Level Regularization}
\label{MLLM:para_reg}
Feature-level distillation preserves prior knowledge by constraining model updates through additional losses, while parameter-level regularization, shown in Figure~\ref{fig:mllm}~(b), directly regularizes model parameters. In MLLMs, some methods argue that only a small subset of parameters is critical during downstream training, with the majority contributing little. Thus, continual learning can focus on selectively protecting these important parameters to reduce forgetting. For instance, Model Tailor~\cite{zhu2024model} defines sensitivity- and salience-based metrics to identify a minimal ``model patch'' crucial for the current task and applies a Hessian-based weight compensation called patch decorator to enhance learning. SPIDER~\cite{huang2024learn} evaluates parameter importance by comparing weight magnitudes of the original model and gradient changes in the fine-tuned model, adaptively fusing weights to preserve generic capabilities while learning new tasks. SEFE~\cite{chen2025sefe} distinguishes between superficial forgetting, addressed by answer style diversification to standardize response formats, and essential forgetting, tackled via RegLoRA, which aggregates regularization masks from prior tasks and optimizes masked parameters toward zero to preserve critical knowledge.
LoRASculpt~\cite{liang2025lorasculpt} applies sparse updates and knowledge-guided regularization to LoRA, removing redundancy and aligning general and specialized knowledge in MLLMs via sparsity and conflict mitigation.
LLaVA-c~\cite{liu2025llava} employs spectral-aware consolidation and unsupervised inquiry regularization to balance task-specific adaptation with general knowledge preservation in MLLMs.

\subsection{Replay-based Approach}
\label{MLLM:replay}
As shown in Figure~\ref{fig:mllm}~(c), replay methods in MLLMs typically store paired image inputs and corresponding instruction data or tokens. For example, VQACL~\cite{zhang2023vqacl} proposes a rehearsal-based method that disentangles sample-specific and sample-invariant features to separate reasoning skills and visual concepts. Sample-specific features are captured via self-attention, while sample-invariant features are learned through dynamically updated prototypes to balance plasticity and stability. SGP~\cite{lei2023symbolic} uses scene graphs as compact visual summaries to replay pseudo scene graph QA triplets; a Symbolic Replay Model captures reasoning and task-specific mappings, while a Unified VQA Model learns from current and replayed data to mitigate forgetting and privacy issues. 
ProtoGroup~\cite{zhang2025multi} clusters related prototypes into groups to form stable invariant representations, selecting memory samples via prototype group correlations.
Lin et al.~\cite{lin2025vlm} leverage memory replay with TF-IDF and K-means clustering to store representative samples, while knowledge distillation guides the model through teacher-student learning to preserve prior knowledge. Additionally, task-specific projection layers are introduced to regularize feature embeddings, minimizing drift across tasks and ensuring continuity in learning, thereby addressing catastrophic forgetting in autonomous driving VQA.  Adapt-$\infty$~\cite{maharana2024adapt} introduces a dynamic data selection framework that clusters samples into pseudo-skill groups via gradient-based representations and adaptively selects high-impact samples using a multi-way scoring mechanism. To maintain efficiency, it permanently prunes semantically redundant data while preserving task balance.  OASIS~\cite{lee2025oasis} adapts sample selection by forward-pass Fisher Information and gradient similarity, dynamically updating selections with exponential moving averages to avoid costly backward passes.

\subsection{Discussions}

For continual learning tasks in Multimodal Large Language Models, approaches leveraging architectural extensions, particularly mixture-of-experts models, continue to dominate. However, although existing multimodal continual learning methods utilize multimodal datasets and architectures, they are primarily designed for general-purpose tasks. In other words, these methods do not differ substantially from continual learning strategies applied to language-only LLMs. Consequently, challenges related to modality-specific forgetting and the influence of inter-modal interactions during continual learning remain largely unexplored and warrant further investigation.

\section{Continual Learning for Vision-Language-Action Models}
\label{Sec:VLA}

\begin{figure}
    \centering
    \includegraphics[width=0.8\linewidth]{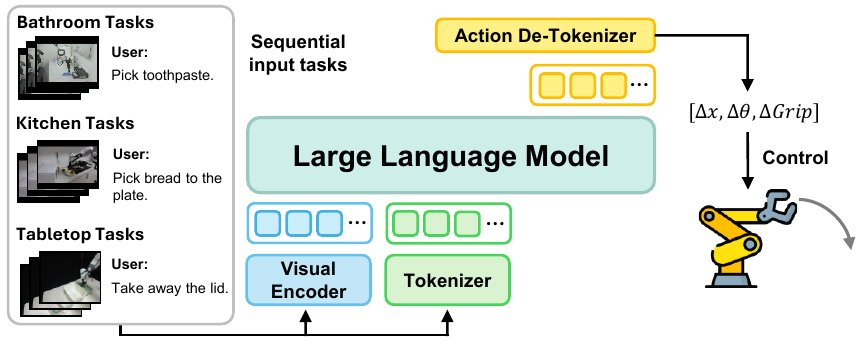}
    \caption{Illustration of a continual learning framework applied to Vision-Language-Action Models.}
    \label{fig:vla}
\end{figure}

Vision-Language-Action~(VLA)~\cite{chi2023diffusion,brohan2023rt,kim2024openvla, zawalski2024robotic} models refer to exploiting large models to predict action and control behaviors of robots~\cite{hansen2022temporal,lopez2025multi} with the aid of a large vision language model. It is of great significance to replace the manipulation of humans in certain risky situations and therefore, improve the level of automation. However, robots are inherently adapting to a particular field and inevitably encounter the problem of poor generalization~\cite{li2025hamster} when dealing with the complex real world. Consequently, continual learning in vision language action~\cite{liu2023libero} aims to equip robots with the ability to adapt and acquire new skills in dynamic and ever-changing real-world scenarios as humans do. It is a specific aspect of lifelong robotic learning~\cite{chen2023fast, gao2021cril, ben2022lifelong, xie2022lifelong, zhang2023dynamics} that employs large multimodal models for robot control and is crucial for robots to accumulate knowledge in novel environments while retaining the ability for known skills to mitigate catastrophic forgetting. Figure~\ref{fig:vla} illustrates the task formulation of visual-language action models within a continual learning framework.

Due to its wide applicability and potential, many articles have investigated the problem from many aspects based on the general strategy. In the following section, we first introduce the problem setup~(Section~\ref{sec:vla-setup}) with common training objectives and benchmarks, then we will categorize vision language action methods into three detailed types: replay-based~(Section~\ref{sec:vla-replay}), architecture-based~(Section~\ref{sec:vla-arch}), and regularization-based~(Section~\ref{sec:vla-reg}) approaches, which will be displayed in separate paragraphs.

\subsection{Problem Setup}
\label{sec:vla-setup}
Lifelong learning~\cite{kirkpatrick2017overcoming, yoo2024exploratory} in vision language action extends the concept of the traditional robot action control framework by requiring one robot to continually acquire and adapt across a sequence of tasks $\{T^1, \dots, T^K\}$ 
over its lifespan while retaining knowledge of previous tasks. This continual learning problem for robots can be formulated as a finite-horizon Markov Decision Process~(MDP): $\mathcal{M} = (\mathcal{S}, \mathcal{A}, \mathcal{T}, H, \mu_0, R)$, where $\mathcal{S}$ represents the state space, $\mathcal{A}$ is the action space, $\mathcal{T} : \mathcal{S} \times \mathcal{A} \rightarrow \mathcal{S}$ is the transition function, $H$ is the maximum horizon for each episode of a task, $\mu_0$ is the initial state distribution, and $R : \mathcal{S} \times \mathcal{A} \rightarrow R$ is the reward function. During the whole training procedure, each task $T^k$ trains the policy $\pi$ from the last task characterized by its own initial state distribution $\mu_0^k$, while $\mathcal{S}, \mathcal{A}, \mathcal{T}$ and $H$ remain unchanged across all tasks.

\noindent
\textbf{Training objectives.} Continual learning in the vision-language-action model sequentially trains one single policy $\pi$ through multiple tasks. For each task $T^k, k\in\{1,\cdots,K\}$, the robot is trained on a dataset of $N$ demonstrations $D^k = \{\tau^k_i\}_{i=1}^{N}$ with a corresponding language description $l^k$ guiding the goal of the robot. Each trajectory $\tau^k_i$ in expert demonstrations with length $L^k < H$ consists of state-action pairs $\{(s_t^k, a_t^k)\}_{t=1}^{L^k}$ sampled from policy $\pi$ conditioned with task $T^k$. The policy is trained with a behavioral cloning loss~\cite{bain1995framework} that aims to mimic the action demonstrations in the dataset:
\begin{align}
    \min_{\pi} \mathcal{L}(\pi) = \frac{1}{K} \sum_{k=1}^{K} \mathbb{E}_{s_t^k, a_t^k \sim D^k} \left[ \sum_{t=1}^{L^k} -\log \pi(a_t^k | s_{\leq t}^k; T^k) \right].
\end{align}
In the scenario of continual learning, the single policy $\pi$ has to continually adapt to specific requirements of sequential tasks $T^k$, while retaining knowledge of gained skills to mitigate forgetting.

\noindent \textbf{Benchmarks.} To evaluate the performance of robots in continual learning, many benchmarks have been constructed in continual learning scenarios from reinforcement learning to vision language action~\cite{wolczyk2021continual, powers2022cora, liu2023libero}. For example, Continual World~\cite{wolczyk2021continual} builds a continual reinforcement learning benchmark on robotic manipulation tasks. It provides a realistic, computationally accessible testbed that highlights the limitations of existing continual learning methods in reinforcement learning settings, which often struggle with forward transfer despite mitigating forgetting. LIBERO~\cite{liu2023libero} builds an imitation learning environment for robots to analyze the continual learning problem of robots from many factors, such as algorithms, architectures, task sequences, and so on, enabling a comprehensive setting for further investigations.


\subsection{Architecture-based Approach}
\label{sec:vla-arch}
Architecture-based continual learning approaches focus on enhancing skill representation and reuse through structural design. One line of work develops skill-aware architectures that distinguish, merge, and update skills~\cite{mete2024quest, wan2024lotus, meng2025preserving} or create modular, extensible frameworks for skill composition~\cite{dong2025optimizing, rauso2024incremental}. For example, LOTUS~\cite{wan2024lotus} incrementally builds a skill library by unsupervisedly segmenting human demonstrations with open-vocabulary vision models, updating skills via clustering, and composing them with a meta-controller for lifelong manipulation. IsCiL~\cite{lee2024incremental} employs a prototype-based memory to learn shareable skills, retrieving relevant prototypes via state embeddings and preserving knowledge with adapters. LEGION~\cite{meng2025preserving} combines Bayesian non-parametric clustering and language-conditioned embeddings, dynamically organizing skills with a Dirichlet Process Mixture Model to scale skill accumulation without performance loss. RLCM~\cite{mendez2022modular} presents a modular lifelong RL framework that dynamically composes policies from reusable neural modules. It combines online module selection and adaptation with offline consolidation via batch RL and experience replay to ensure long-term skill retention and generalization.
Jia et al.~\cite{jia2025hierarchical} propose hierarchical routing with cross-modal clustering and SVD-based incremental LoRA, preserving principal components and orthogonally training residuals to reduce catastrophic forgetting in embodied continual learning.
DRAE~\cite{long2025drae} integrates dynamic MoE routing with parameterized RAG and hierarchical RL to enable lifelong learning while mitigating catastrophic forgetting through spectral-aware consolidation and DPMM-based knowledge retention.

Another line leverages Large Language Models (LLMs) for high-level task guidance in robotics, such as planning~\cite{parakh2024lifelong} and skill demonstration~\cite{tziafas2024lifelong, barmann2024incremental}. These methods utilize pre-trained LLMs as knowledge bases, retrieving task-relevant information via textual prompts to aid skill acquisition and transfer. Thanks to their data efficiency and fast inference, LLMs show great promise in diverse robotic continual learning scenarios.

\begin{table*}[!t]
    \centering
    \caption{Summary of continual learning methods for Vision-Language-Action Models. Taxonomy: \textcolor{red}{\ding{52}} indicates the primary solution, and \ding{52} indicates the supporting method. \textbf{Tag}: \encircle[fill=lightcoral, text=white]{I} = \underline{I}mage, \encircle[fill=brightlavender, text=white]{T} = \underline{T}ext, \encircle[fill=harvestgold, text=white]{S} = \underline{S}ensor.}
    \resizebox{0.98\textwidth}{!}{
    \begin{tabular}{lcccllccc}
    \toprule
       \multirow{2}{*}{\textbf{Methods}} & \multicolumn{3}{c}{\textbf{Taxonomy}} & \multirow{2}{*}{\textbf{Benchmark / Dataset}} & \multirow{2}{*}{\textbf{Backbone}} & \multirow{2}{*}{\textbf{Modality}} & \multirow{2}{*}{\textbf{Environmental Setting}} & \multirow{2}{*}{\textbf{Code}} \\ 
       \cmidrule{2-4}
       & \textbf{Arch.} & \textbf{Reg.} & \textbf{Rep.} \\ \midrule

       \rowcolor[rgb]{ .949,  .949,  .949} LIBERO~\cite{liu2023libero} & \textcolor{red}{\ding{52}} & \ding{52} & \ding{52} & LIBERO & ViT-T, ResNet-T, BERT & \encircle[fill=lightcoral, text=white]{I} \encircle[fill=brightlavender, text=white]{T} & Simulated & \href{https://github.com/Lifelong-Robot-Learning/LIBERO}{Link} \\

       LOTUS~\cite{wan2024lotus} & \textcolor{red}{\ding{52}} & \ding{52} & \ding{52} & LIBERO & DINOv2, BERT &\encircle[fill=lightcoral, text=white]{I} \encircle[fill=brightlavender, text=white]{T} & Realistic \& Simulated & \href{https://github.com/UT-Austin-RPL/Lotus}{Link} \\

       \rowcolor[rgb]{ .949,  .949,  .949} QueST~\cite{mete2024quest} & \textcolor{red}{\ding{52}} &\ding{52} & & LIBERO, MetaWorld & ResNet18, CLIP & \encircle[fill=lightcoral, text=white]{I} \encircle[fill=brightlavender, text=white]{T} & Simulated & \href{https://github.com/pairlab/QueST}{Link} \\

       LEGION~\cite{meng2025preserving} & \textcolor{red}{\ding{52}} & \ding{52}& \ding{52}& LEGION & RoBERTa & \encircle[fill=lightcoral, text=white]{I} \encircle[fill=brightlavender, text=white]{T} & Realistic \& Simulated & \href{https://github.com/Ghiara/LEGION}{Link} \\

       \rowcolor[rgb]{ .949,  .949,  .949} Jia et al.~\cite{jia2025hierarchical} & \textcolor{red}{\ding{52}} & \ding{52}& & ALFRED & LLAMA-2-7B, CLIP & \encircle[fill=lightcoral, text=white]{I} \encircle[fill=brightlavender, text=white]{T} & Simulated & - \\

       DRAE~\cite{long2025drae} & \textcolor{red}{\ding{52}} &  & \ding{52} & MimicGen, NAVSIM & ResNet & \encircle[fill=lightcoral, text=white]{I} \encircle[fill=brightlavender, text=white]{T} & Simulated & - \\

        CAMA~\cite{kim2024online} &  &  \textcolor{red}{\ding{52}}& & ALFRED &  ResNet, Bi-LSTM & \encircle[fill=lightcoral, text=white]{I} \encircle[fill=brightlavender, text=white]{T} & Simulated & \href{https://github.com/snumprlab/cl-alfred}{Link}\\

        \rowcolor[rgb]{ .949,  .949,  .949}   M2Distill~\cite{roy2024m2distill} &  &\textcolor{red}{\ding{52}} & & LIBERO &  ResNet, BERT & \encircle[fill=lightcoral, text=white]{I} \encircle[fill=brightlavender, text=white]{T} \encircle[fill=harvestgold, text=white]{S}& Simulated & - \\

        iManip~\cite{zheng2025imanip} & {\ding{52}} & \ding{52}& \textcolor{red}{\ding{52}} & RLBench & CLIP, U-Net & \encircle[fill=lightcoral, text=white]{I} \encircle[fill=brightlavender, text=white]{T} & Realistic \& Simulated & - \\

        \rowcolor[rgb]{ .949,  .949,  .949} RWLA~\cite{yang2024task} &  & & \textcolor{red}{\ding{52}}& LIBERO & R3M, SS & \encircle[fill=lightcoral, text=white]{I} \encircle[fill=brightlavender, text=white]{T} & Simulated & - \\       
    \bottomrule
    \end{tabular}}
    \vspace{-10pt}
    \label{tab:vla}
\end{table*}

\subsection{Regularization-based Approach}
\label{sec:vla-reg}
Regularization methods for continual learning usually try to keep important parameters or features unchanged and impose feature-level or parameter-level regularization during training, therefore maintaining the ability for past skills~\cite{haldar2023polytask, kim2024online, roy2024m2distill, lee2024incremental}. CAMA~\cite{kim2024online} introduces Confidence-Aware Moving Average, a method that dynamically updates stored knowledge~(logits) using confidence scores to balance past and current information without requiring task boundaries. PolyTask~\cite{haldar2023polytask} proposes a method combining task-specific policy training with behavior distillation, which consolidates individual policies into a single model using offline data. With demonstration-guided reinforcement learning and distilling policy outputs, it handles continual actions and lifelong learning by avoiding interference between tasks. M2Distill~\cite{roy2024m2distill} proposes a multi-modal distillation framework that preserves consistency across vision, language, and action modalities between incremental learning steps. It exploits: (1) multi-modal feature distillation to constrain latent space drift; and (2) policy distillation to maintain alignment of Gaussian mixture model~(GMM) action distributions. The approach ensures skill retention without storing prior data, advancing lifelong learning for real-world robotic applications. ELIRL~\cite{mendez2018lifelong} introduces Lifelong Inverse Reinforcement Learning~(Lifelong IRL), addressing the challenge of learning multiple sequential tasks and proposing Efficient Lifelong IRL to enable knowledge transfer across tasks, improving performance while minimizing user-provided data.

\subsection{Replay-based Approach}
\label{sec:vla-replay}
Due to the spatial and temporal complexity of robotic manipulation in various and dynamic scenarios, it has become common practice to store part of previous samples and replay them~\cite{bang2021rainbow, rolnick2019experience} during the training procedure to retain the knowledge of previous tasks~\cite{zheng2025imanip, meng2025preserving, dong2025optimizing, yang2024task}.
iManip~\cite{zheng2025imanip} addresses the challenge of catastrophic forgetting in robotic manipulation. Given that traditional methods fail to retain previously learned skills, ignoring temporal complexity and action complexity, the introduced framework integrates a temporal replay strategy to preserve task continuity and an extendable PerceiverIO~\cite{jaegle2021perceiver} architecture with adaptable action prompts. Experimental results highlight the efficiency, scalability, and robustness across complex, long-horizon tasks. Decision-RWKV~\cite{dong2025optimizing} exploits a novel sequence modeling framework that integrates the efficiency of the Receptance Weighted Key Value~(RWKV) with experience replay mechanisms. It leverages the linear complexity and memory retention capabilities of RWKV to enable efficient lifelong learning across diverse manipulation tasks, and the experience replay combines a replay buffer for historical experiences reuse while acquiring new skills. RWLA~\cite{yang2024task} combines retrieval-based local adaptation with a selective weighting mechanism to restore forgotten skills. The method retrieves relevant past demonstrations from a memory buffer, identifies critical failure-prone segments, and prioritizes these segments during local fine-tuning, which allows robots to adapt continually while maintaining performance across diverse manipulation tasks.

\subsection{Discussions}

Continual learning in Vision-Language-Action models increasingly adopts hybrid strategies that integrate architectural expansion, replay mechanisms, and regularization constraints. While early methods often apply these components independently, recent advances highlight the importance of balancing them to achieve effective continual learning. Specifically, architectural modifications enable task-specific adaptability, replay mechanisms help preserve cross-task consistency, and regularization mitigates catastrophic interference across modalities. However, key challenges remain, particularly in dynamically coordinating these strategies over time and addressing the tendency of action policy components to forget more rapidly than vision-language alignment modules.

\section{Continual Learning for Diffusion Models}
\label{Sec:Diff}

Latent Diffusion Models~(LDMs) have emerged as a powerful tool in artificial intelligence, capable of generating photorealistic and semantically consistent images from user-defined textual prompts. Building on this capability, various Personalized Diffusion Models~(PDMs) have been developed to generate images tailored to specific user input data~(also known as concepts) and textual prompts, significantly enhancing interactivity. However, as shown in Figure~\ref{fig:dm}, when user inputs and customization requirements arrive sequentially, these methods tend to forget previously learned concepts. Thus, overcoming forgetting in diffusion models under continual customization scenarios remains a key challenge.

In the following section, we first introduce the problem setup~(Section \ref{Diff:setup}), including training objectives and commonly used benchmarks. Then, we summarize existing research by categorizing it into architecture-based~(Section \ref{Diff:arch}) and regularization-based~(Section \ref{Diff:reg}) approaches.

\subsection{Problem Setup}
\label{Diff:setup}
\textbf{Training objectives.} Diffusion models are probabilistic generative models that learn data distributions by iteratively denoising Gaussian noise. Given an initial noise map $\varepsilon \sim \mathcal{N}(\mathbf{0}, \mathbf{I})$, timestep $t \sim \text{Uniform}({1, \dots, T})$, and user prompt $p$, an image is generated as $x = \mathcal{D}(\epsilon_{\theta}(c, t))$, where $c = \Gamma(p)$ is the text embedding from encoder $\Gamma$. The denoising model $\epsilon_\theta$ is trained with:
\begin{equation}
    \mathcal{L}_{\text{DM}}(\theta)=\mathbb{E}_{z,\varepsilon,c,t}[||\varepsilon-\epsilon_{\theta}(z_t, c, t)||^2],
\label{eq:diff}
\end{equation}
\noindent
where $z_t$ is the noisy image at timestep $t$, and $\theta$ denotes the parameters of the denoising model.

To better reflect user intent, PDMs enhance generation by incorporating a few user images alongside text prompts to bind novel concepts (e.g., a person or object) to a unique identifier while retaining generalization. DreamBooth~\cite{ruiz2023dreambooth} introduces a class-specific prior preservation loss to prevent overfitting and preserve base class knowledge:
\begin{equation}
    \mathcal{L}_{\text{PDM}}(\theta) = \mathbb{E}_{z,\varepsilon,c,t}\left[\left\|\varepsilon - \epsilon_{\theta}(z_t, c, t)\right\|^2\right]
    + \lambda \, \mathbb{E}_{z^p,\varepsilon,c^p,t}\left[\left\|\varepsilon - \epsilon_{\theta}(z_{t}^p, c^p, t)\right\|^2\right],
\end{equation}
\noindent
where $z_t$ denotes the latent code of an image with the new concept, and $z_t^p$ corresponds to a class-specific image used for prior preservation. $c$ and $c^p$ are the respective conditioning vectors; for instance, $c$ may be ``a photo of [$\mathcal{V}^*$] dog'', while $c^p$ is ``a photo of dog''. The special token [$\mathcal{V}^*$] is a unique identifier representing the user-specific concept learned from the provided images. $\lambda$ controls the regularization strength.

\begin{figure}[t]
    \centering
    \includegraphics[width=0.98\linewidth]{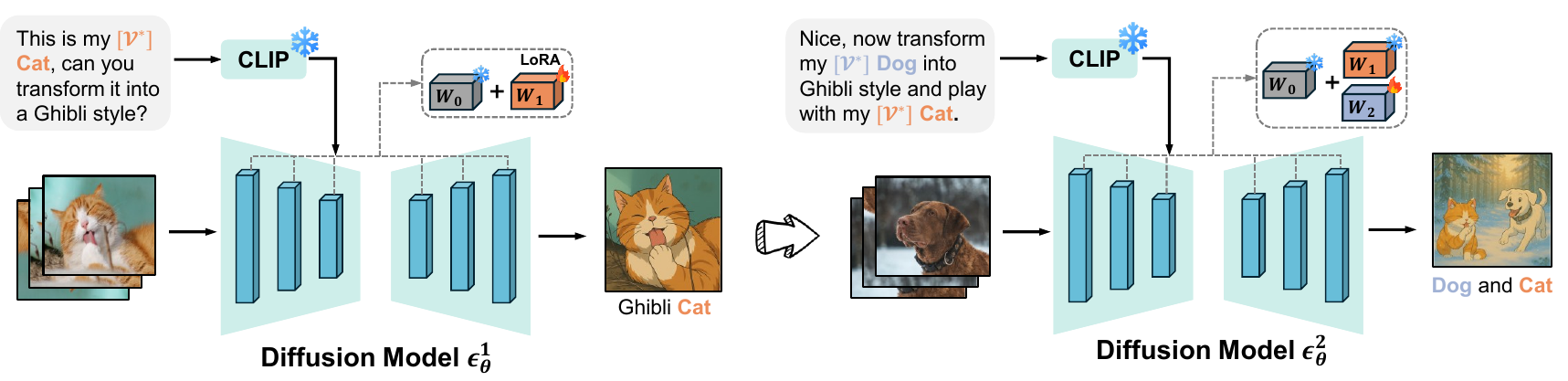}
    \caption{Illustration of a continual learning framework applied to Diffusion Models.}
    \label{fig:dm}
    \vspace{-10pt}
\end{figure}

Continual learning in diffusion models extends beyond single-concept personalization by incrementally acquiring new concepts without retraining on prior data or suffering catastrophic forgetting. As illustrated in Figure~\ref{fig:dm}, let $\mathcal{T} = {\{\mathcal{T}^k}\}_{k=1}^K$ be a sequence of $K$ generation tasks with datasets ${\{\mathcal{D}^k}\}_{k=1}^K$. Each task $\mathcal{T}^k$ contains $n_k$ image-prompt pairs $(x_i^k, c_i^k)$, where $n_k$ is typically 3-5. The model learns the $k$-th concept via the following loss:
\begin{equation}
    \mathcal{L}_{\text{CDM}}(\theta) = \mathbb{E}_{z^k,\varepsilon,c^k,t}\left[\left\|\varepsilon - \epsilon_{\theta}^k(z_{t}^k, c^k, t)\right\|^2\right] 
    + \lambda \, \mathbb{E}_{z^{k,p},\varepsilon,c^{k,p},t}\left[\left\|\varepsilon - \epsilon_{\theta}^k(z_{t}^{k,p}, c^{k,p}, t)\right\|^2\right],
\end{equation}
\noindent
where $(z_t^k, c^k)$ denote the image latent code at timestep $t$ and text embeddings corresponding to the $k$-th generation task, and $(z_{t}^{k,p}, c^{k,p})$ represent the prior pairs generated by the pre-trained model for the same task.

\noindent
\textbf{Backbone.} Stable Diffusion~(SD-1.5)~\cite{rombach2022high} is a high-resolution image synthesis model based on LDMs, offering a powerful and efficient generative framework. It operates by encoding images into a lower-dimensional latent space using a pre-trained autoencoder, significantly reducing computational costs during training and inference. The model then performs denoising in this latent space with a U-Net backbone guided by a text encoder, such as CLIP, enabling high-fidelity image generation conditioned on text prompts. In continual personalized diffusion settings, LoRA-based fine-tuning is typically applied to the U-Net backbone to efficiently adapt to limited training data.

\noindent
\textbf{Benchmark.} CLoG~\cite{zhang2024clog} proposes two benchmarks: Label-conditioned and Concept-conditional, to evaluate the effectiveness of continual learning approaches in the image generation field. The label-conditioned setting requires the model to learn a sequence of generative tasks conditioned on label indices, while the concept-conditional setting focuses on synthesizing a series of personalized concepts, denoted as $\mathcal{V}^*_k$, where $k$ indicates the $k$-th concept.

\subsection{Architecture-based Approach}
\label{Diff:arch}

C-LoRA~\cite{smith2023continual} introduces a continual concept learning setting within the LDM framework, where each user-defined concept is assigned a dedicated LoRA module injected into the self-attention keys and values. During training, each concept is represented by a unique placeholder token~(\eg~[$\mathcal{V}^*$]) inserted into the prompt to associate the visual concept with a textual identifier. To mitigate catastrophic forgetting, C-LoRA freezes the LoRA parameters associated with previously learned concepts during new concept learning. Furthermore, it incorporates a self-regularization loss to penalize updates to weight regions that were modified by earlier concepts, thus preserving prior knowledge.

\subsection{Regularization-based Approach}
\label{Diff:reg}

MuseumMaker~\cite{liu2024museummaker} retains previously learned styles while integrating new custom ones. To prevent overfitting and preserve original diffusion concepts, it introduces a style distillation loss that captures disentangled style representations. It also proposes a dual regularization scheme for LoRA training, constraining optimization from both weight and feature perspectives to balance knowledge integration.
LFS-Diffusion~\cite{song2024towards} adopts a data-free distillation strategy across diffusion time steps to curb overfitting to new concepts and preserve generative capacity. In inference, an in-context generation mechanism enhances the retention of prior knowledge.
CIDM~\cite{dong2024continually} proposes a concept consolidation loss using learnable layer-wise concept tokens and an orthogonal subspace regularizer to separate task-specific and shared knowledge. It further introduces elastic weight aggregation for merging concept-specific LoRA weights and a context-controllable synthesis strategy that enhances regional features via layer-wise text embeddings and region-aware noise.
ConceptGuard~\cite{guo2025conceptguard} mitigates forgetting through shift embeddings for dynamic updates, concept-binding prompts for composition, and memory preservation regularization to limit parameter drift. A priority queue manages concept replay based on importance and recency.
Jha et al.~\cite{jha2025mining} leverage diffusion classifier scores as implicit class-conditional density estimators to guide continual adaptation. By regularizing both parameter and function spaces, their method enables diffusion models to adapt to evolving user concepts while minimizing forgetting of prior knowledge.

\subsection{Replay-based Approach}
\label{Diff:replay}

Since each stage in the continual personalized diffusion setting typically involves only a limited number of training samples (\eg~3-5 images), a feature replay strategy is commonly adopted. Specifically,  $\text{L}^2\text{DM}$~\cite{sun2024create} addresses the continual concept learning challenge by introducing the Rainbow-Memory Bank Strategy, which consists of a short-term memory bank for encoding current task features, a long-term memory bank for preserving historical knowledge, and a scoring function to balance knowledge retention and transfer. To further mitigate the forgetting of personalized details, $\text{L}^2\text{DM}$ proposes Elastic Concept Distillation, which enhances model performance by distilling knowledge between the current model and the previous task model, using features retrieved from the short-term memory bank as input to both.

\subsection{Discussions}

In this section, we discuss continual learning in diffusion models, with a particular focus on continual personalized generation. Due to the limited training data per task and the relatively low variance across tasks in this setting, regularization-based approaches remain predominant. Nevertheless, we argue that a major limitation in this area lies in the absence of standardized evaluation benchmarks and robust quantitative metrics, which are essential for fairly assessing different methods. Furthermore, exploring broader continual learning scenarios within diffusion models represents a promising and underexplored research direction.

\section{Future Directions}
\label{sec:future}

Despite significant advances in continual learning for generative models, critical challenges persist across multiple dimensions. In this section, we further identify promising research directions at the intersection of continual learning and generative modeling, aiming to stimulate further progress in the field.

\noindent \textbf{Efficient Continual Learning Mechanisms.} Current mainstream continual learning approaches often rely on model expansion to retain task-specific knowledge, which leads to substantial storage and computational overhead during deployment. These challenges are particularly pronounced for large-scale generative models, where the large number of parameters causes latency and resource limitations. Consequently, there is increasing interest in enhancing efficiency through techniques such as model compression~\cite{zhu2024survey, li2024tokenpacker, chen2025recoverable}, pruning~\cite{cheng2024survey, wu2023ppt, an2024fluctuation, zeng2025token, castells2024ld, kong2025token}, and quantization~\cite{liu2024spinquant, zhao2024atom, lin2024awq}.
In addition to structural optimization, recent progress in dataset compression and distillation~\cite{wang2018dataset, liu2022dataset, ma2025towards, wang2025dataset} highlights the potential of compressing replay data as a promising direction. A key idea is to encode as much task-relevant information as possible from previous data or features into a small number of representative tokens. This strategy not only helps mitigate catastrophic forgetting but also reduces the risk of data privacy leakage by avoiding the storage of raw data.
Overall, integrating continual learning with efficient model and data compression techniques provides a compelling path toward improving the practicality and scalability of generative models in real-world applications.

\noindent \textbf{Learning Paradigm for Continual Learning.} While conventional continual learning methods predominantly rely on supervised fine-tuning~(SFT) for incremental knowledge acquisition, emerging reinforcement learning~(RL)-based training paradigms, particularly exemplified by GRPO~\cite{shao2024deepseekmath}, have demonstrated stronger generalization capabilities across a variety of model architectures. This performance advantage has been consistently observed not only in large language models~\cite{guo2025deepseek, li2025system, yu2025dapo} but also in multimodal large language models~\cite{liu2025visual, huang2025vision, shen2025vlm, peng2025lmm} and vision-language-action models~\cite{guo2025improving, zhang2024grape}, suggesting that RL-based methods may serve as a more robust alternative to traditional SFT strategies in continual learning settings. Recent studies~\cite{chu2025sft} have revealed a fundamental difference between the two paradigms: SFT tends to overfit specific training distributions through pattern memorization, whereas RL emphasizes structural comprehension of task objectives, enabling better generalization to novel scenarios. This distinction provides strong motivation for exploring continual learning within an RL framework, which may naturally support long-term adaptation, task abstraction, and reinforcement-driven retention strategies.

\noindent \textbf{Self-generation for Continual Learning.} A core challenge in continual learning lies in the inaccessibility of previous task data. In discriminative models, a widely adopted solution involves training auxiliary generative models~(\eg, GANs~\cite{goodfellow2014generative} or diffusion models) to synthesize pseudo-samples of prior tasks, thereby mitigating catastrophic forgetting~\cite{shin2017continual, wu2018memory, cong2020gan, meng2024diffclass}. While effective, this approach introduces significant computational overhead and may suffer from a distributional mismatch between generated and original data. In contrast, generative models offer a more integrated solution by leveraging their inherent capacity to reconstruct and regenerate data. These models can autonomously replay past task distributions during the learning of new tasks, without the need for separate generative modules. This capability offers two major benefits: reducing additional architectural complexity and achieving better alignment with the original data distribution. As a result, self-generation in generative models presents a promising direction for building more efficient, scalable, and distributionally consistent continual learning systems.

\noindent \textbf{Continual Learning in Large Scale Models.} Most existing studies on continual learning for generative models have focused on small- to medium-scale architectures (e.g., models with up to 7B and 13B parameters). However, emerging evidence~\cite{bi2024deepseek, zhang2024scaling} suggests that continual learning behaviors can differ substantially as the model scale increases. This observation highlights two critical research directions. First, there is a pressing need to establish specialized evaluation benchmarks that mitigate information leakage risks~\cite{kim2023learnability}, particularly given the large and often proprietary datasets used to train large-scale models. Second, it is essential to explore parameter-efficient knowledge integration strategies—such as Chain-of-Thought reasoning~(CoT)\cite{wei2022chain, zhang2023multimodal}, In-Context Learning~(ICL)\cite{dong2022survey, momeni2024context, xu2024stress}, Agents~\cite{xi2025rise, xie2024large, zheng2025lifelongagentbench}, and Retrieval-Augmented Generation~(RAG)~\cite{lewis2020retrieval, gao2023retrieval}—as conventional fine-tuning becomes increasingly infeasible for models with extensive parameter counts.

\noindent \textbf{Continual Learning for Multimodal Generation Beyond Vision and Language.} While most existing research in continual generative learning focuses on vision-language tasks, real-world multimodal systems are increasingly incorporating richer modalities such as audio~\cite{zhang2023speechgpt}, 3D scenes~\cite{hong20233d}, and motion data~\cite{jiang2023motiongpt}. In such settings, modality-continual learning becomes a critical yet underexplored challenge. For example, a pre-trained MLLM may be required to incorporate speech input or generate spoken responses. Without appropriate continual learning mechanisms, learning the new audio modality could result in severe forgetting of previously acquired vision-language capabilities. Similarly, in VLA models deployed in robotics or embodied agents, the integration of new sensor modalities (\eg~depth and haptic signals) necessitates stable expansion of generative capacity without overwriting existing knowledge. These scenarios highlight the need for continual learning methods capable of integrating new modalities while preserving alignment and generative quality across previously learned ones.

\noindent \textbf{Toward Unified Optimization Frameworks for Continual Learning in Generative Models.} Although generative models vary widely in data modalities and task objectives, increasing architectural convergence is fostering the development of unified continual learning frameworks. For example, both multimodal large language models and vision-language-action models frequently rely on large language models to enable cross-modal reasoning. Furthermore, recent attempts to integrate diffusion models with MLLMs~\cite{xie2024show, ge2024seed, wu2024next} underscore a broader trend toward unifying generation and understanding within generative AI. These shared characteristics suggest a promising direction: designing continual learning strategies that generalize across generative model families. Unifying optimization across modalities and architectures may mark a critical step toward deploying generative AI in complex, dynamic real-world scenarios.

\section{Conclusion}
\label{sec:conclusion}

In this survey, we have reviewed recent advances in continual learning for generative models, providing a systematic overview across four major categories: Large Language Models, Multimodal Large Language Models, Vision-Language-Action Models, and Diffusion Models. We further unify these models by identifying shared methodological dimensions that underpin their continual learning paradigms. Specifically, we formalize the continual learning problem in each domain, introduce relevant datasets and model architectures, and categorize representative approaches within each subfield. We hope that this comprehensive and in-depth survey, along with the accompanying real-time updated \href{https://github.com/Ghy0501/Awesome-Continual-Learning-in-Generative-Models}{GitHub repository}, will not only consolidate the current understanding of continual learning in generative modeling but also serve as a valuable reference and catalyst for future research.

\begin{acks}
This work was supported by the National Science and Technology Major Project (2022ZD0116500), National Natural Science Foundation of China (62222609, 62320106010), CAS Project for Young Scientists in Basic Research (YSBR-083), Major Science and Technology Plan Project on the Future Industry Fields of Xiamen City (3502Z20241027), Unveiling and Leading Projects of Xiamen (3502Z20241011) and the InnoHK program. 
\end{acks}

\bibliographystyle{ACM-Reference-Format}
\bibliography{main}


\end{document}